\pdfoutput=1
\documentclass[11pt]{article}

\usepackage[final]{coling}
\usepackage{hyperref}

\usepackage{array}
\usepackage{wrapfig}
\usepackage{multirow}
\usepackage{xcolor}
\usepackage[amssymb]{SIunits}
\usepackage{amssymb}
\usepackage{graphicx}
\usepackage{tabularx}
\usepackage{pifont}
\usepackage{times}
\usepackage{latexsym}
\usepackage{booktabs}
\usepackage{algorithm}
\usepackage{algpseudocode}
\usepackage{float}

\usepackage{amsmath}
\usepackage[T1]{fontenc}

\usepackage[utf8]{inputenc}

\usepackage{microtype}

\usepackage{inconsolata}
\usepackage{SIunits}

\usepackage{graphicx}
\newcommand{\stdvu}[1]{\scriptsize{\color{darkgray}(#1)}}

\title{MMAC-Copilot: Multi-modal Agent Collaboration Copilot}

\author{
\textmd{Zirui Song}$^{1,2 \ast}$
\quad
\text{Yaohang Li}$^2$\thanks{Equal contributions.}
\quad
\text{Meng Fang}$^{4*}$
\quad
\text{Yanda Li}$^2$
\quad
\text{Zhenhao Chen}$^{1}$ \\
\quad
\text{Yuan Huang}$^{3}$
\quad
\text{Zhecheng Shi}$^{3}$
\quad
\text{Xiuying Chen}$^{1}$
\quad
\text{Ling Chen}$^{2}$ \\
\\[0.1125cm]
\normalsize 
$ ^1$ Mohamed bin Zayed University of Artificial Intelligence 
\normalsize \\
$ ^2 $ University of Technology Sydney \\
$ ^3 $ Northeastern University \\
$ ^4 $ University of Liverpool \\
}

\begin{document}
\maketitle
\begin{abstract}
%Autonomous agents that interact with PC applications are often limited by their singular mode of interaction with real-world environments, restricting their versatility. 
Large language model agents that interact with PC applications often face limitations due to their singular mode of interaction with real-world environments, leading to restricted versatility and frequent hallucinations. 
To address this, we propose the Multi-Modal Agent Collaboration framework (MMAC-Copilot), a framework utilizes the collective expertise of diverse agents to enhance interaction ability with application.
The framework introduces a team collaboration chain, enabling each participating agent to contribute insights based on their specific domain knowledge, effectively reducing the hallucination associated with knowledge domain gaps. We evaluate MMAC-Copilot using the GAIA benchmark and our newly introduced Visual Interaction Benchmark (VIBench). MMAC-Copilot achieved exceptional performance on GAIA, with an average improvement of 6.8\% over existing leading systems.
VIBench focuses on non-API-interactable applications across various domains, including 3D gaming, recreation, and office scenarios.  It also demonstrated remarkable capability on VIBench. We hope this work can inspire in this field and provide a more comprehensive assessment of Autonomous agents. The anonymous Github is available at \href{https://anonymous.4open.science/r/ComputerAgentWithVision-3C12}{Anonymous Github}

\end{abstract}

\section{Introduction}

Large Language Models (LLMs) \cite{achiam2023gpt, anil2023palm, anthropic2023claude2,driess2023palme,ouyang2022training,touvron2023llama2,touvron2023llama} has demonstrated unprecedented capabilities to tackle complex tasks that require human-like reasoning, decision-making, and collaborative efforts \cite{ding2023everything,hao2023reasoning,park2023generative,qian2023communicative,xi2023rise}. Soon afterwards, Large Vision Models (LVMs) \cite{cai2023benchlmm,li2024llava,liu2023improved,liu2024visual,luo2023wizardcoder,zhou2023lima} expand beyond traditional text-based processing by integrating a visual dimension into LLMs. The applications of LVMs have reached diverse domains, with one notable example being their pivotal usage in autonomous agents 
\cite{wu2024oscopilot}. 

%Usually, autonomous agents require a unified interaction interface and a high degree of generalization to adapt to the challenges posed by the numerous different applications within the operating system. Although previous works \cite{tan2024towards,wu2024oscopilot,zhang2024ufo} that utilize scripts and the planning abilities of LLMs have demonstrated success in some applications there are still some challenges: First, the acquisition of real-world environmental information relies on a single text modality which lead to a limited range of applications. Second, the hallucinations caused by the knowledge domain gap could lead to incorrect planning by virtual agents, resulting in extremely severe consequences.

Applying large language models (LLMs) in operating systems faces two primary challenges. First, their ability to acquire real-world environmental information is constrained by reliance on a single text modality, limiting the breadth of applications they can effectively manage. Second, hallucinations stemming from knowledge domain gaps can lead to flawed planning by LLM agents, with potentially severe consequences. Agents typically require both a unified interaction interface and a high level of generalization to adapt to the diverse and complex range of applications within an operating system. While previous works \cite{tan2024towards,wu2024oscopilot,zhang2024ufo} have leveraged scripting and LLMs' planning abilities to achieve success in certain scenarios, these challenges persist and must be addressed for broader applicability.

%To engage with the above challenges, we introduce MMAC-Copilot, a framework designed to promote information exchange and collaboration among multi-modal agents. 

\begin{figure*}[h]
    \centering
    \includegraphics[width=1\textwidth,height=0.4\textheight]{ 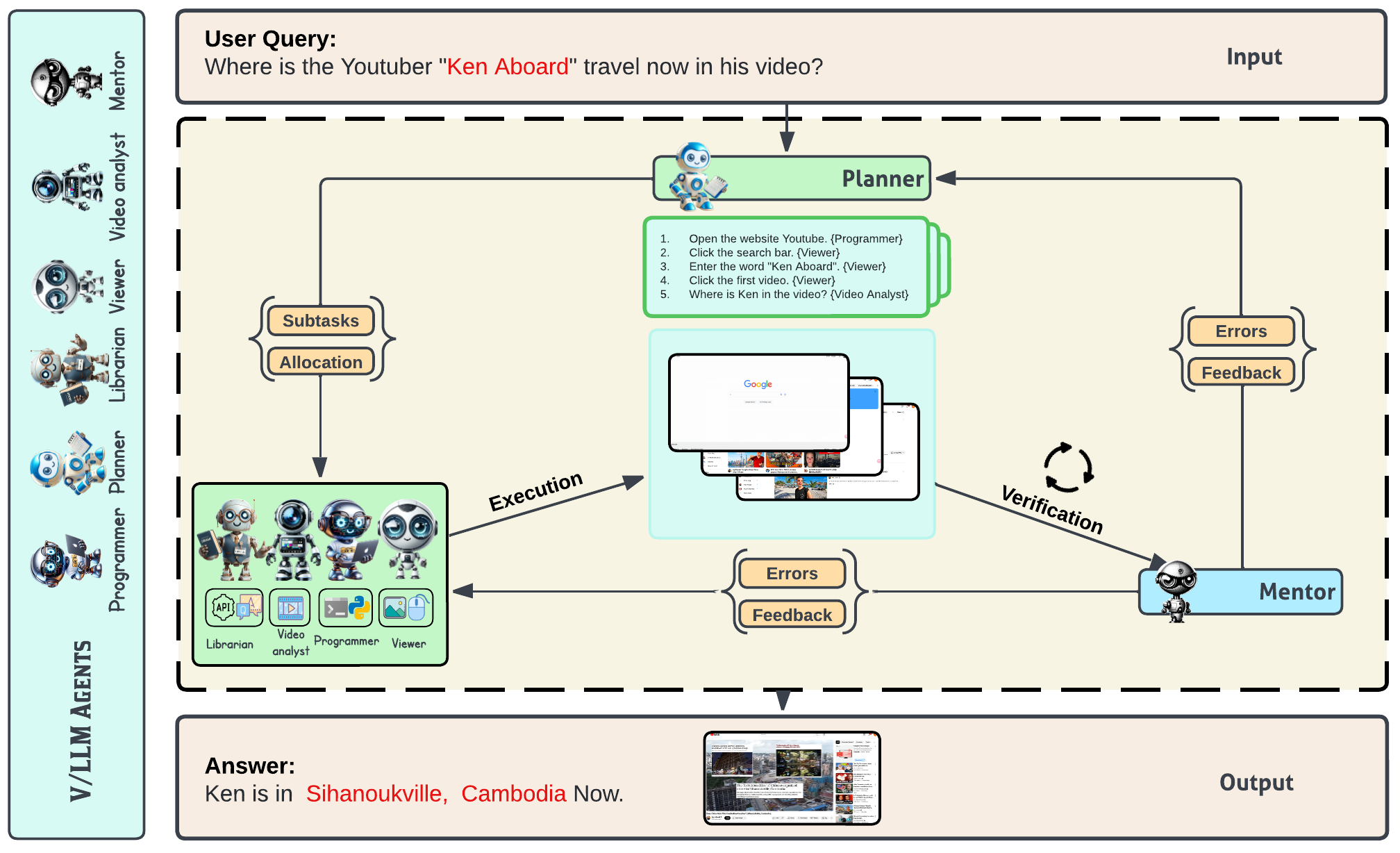}
\caption{An overview of MMAC-Copilot framework.}
    \label{fig:framework}
\end{figure*}

To address the aforementioned challenges, we present MMAC-Copilot, a framework designed to enhance information exchange and collaboration among multi-modal agents. Central to this framework is the team collaboration chain, where agents—each driven by an LLM—adapt the initial plan based on their specific domain expertise. 
When a user query is received, the Planner formulates an initial plan based on the query's requirements. The plan is then iteratively refined through collaboration with the other agents, ensuring that insights from each agent's domain expertise continuously adapt and improve the execution process.
As illustrated in Figure \ref{fig:framework}, MMAC-Copilot integrates five key agents:
\begin{itemize}
\item Planner: Specializes in strategic planning, resource allocation, and ensuring that all agents' efforts are aligned toward achieving the target.
\item Librarian: Excels in question-answering tasks and API-based information retrieval.
\item Programmer: Handles coding tasks, with expertise in executing Bash commands and Python scripts.
\item Viewer: Focuses on processing visual information, understanding image content, and performing click-based interactions.
\item Video Analyst: Analyzes video content to extract key events.
\end{itemize}
In addition to these core agents, an auxiliary agent, the Mentor, supervises the system's interactions with the environment, reviewing the user interface after each interaction to confirm whether the desired outcomes have been achieved.

We evaluate MMAC-Copilot using two benchmarks: GAIA \cite{mialon2023gaia} and our newly introduced Visual Interaction Benchmark (VIBench). GAIA is designed to assess general artificial intelligence assistants, and MMAC-Copilot achieved scores of 45.16, 20.75, and 6.12 on level-1, level-2, and level-3 tasks (as detailed in Appendix \ref{sec:appendix:GAIA}), outperforming previous methods by 6.8\%. VIBench, developed specifically for this study, evaluates performance in non-API-interactable applications across domains such as 3D gaming, recreation, and office environments. The results demonstrate MMAC-Copilot's exceptional adaptability and precision in handling complex interfaces.

% 为了尽可能融合更多的模态信息 我们为agent拼接了 不同的模态工具包，赋予其多模态的分析能力。

% \yd{It seems that there is some overlap in the functional descriptions of planner and mentor? Additionally, regarding the text-only challenge, it seems that only the Viewer and Video Analyst components are related to multimodal interaction. Is the introduction of the remaining components relevant in this context?}

%In resolving the second dilemma, we introduce the team collaboration chain, allowing participating agents driven by LLM to adapt the initial plan crafted based on their domain expertise. 

%\yd{The distinction between the roles of Mentor and Planner doesn't seem clear enough.}

In summary, our main contributions are as follow:

First, we propose MMAC-Copilot, a collaborative framework involving specialized agents designed to navigate and perform general application tasks. MMAC-Copilot enhances autonomous agents' ability to interact with user interfaces when processing tasks.
% By leveraging the collaboration of modal agents, MMAC-copilot can handle a diverse range of tasks with enhanced accuracy.

Next, to mitigate the incorrect planning issue caused by LLMs' hallucination, we introduce the team collaboration chain which allows participating agents to adjust the initial plan based on their domain expertise into MMAC-Copilot. It has demonstrated adaptability in handling a variety of case.  

Finally, we develop the Visual Interaction Benchmark (VIBench), specifically designed to assess the system's performance in non-API-interactable applications across diverse domains such as 3D gaming, recreation, and office scenarios. VIBench complements existing benchmarks by focusing on the agents' ability to interact with visually complex user interfaces.

% \yd{If the mention of the OS in the abstract is not revised, the introduction should also include related content to ensure consistency}

% \mf{Our contributions xx}

\section{Related work}
In this section, we review some related works, with a focus on the domains of LLM-based agents and autonomous agents.

\subsection{LLM-based agents}

The sunrise field of LLM-based agents, highlighted by projects such as AutoGPT \cite{autogpt} and TaskWeaver \cite{qiao2023taskweaver}, marks a significant milestone in the field of LLM research. These agents adopt a human-like approach to problem-solving by decomposing complex tasks into manageable subtasks. The capability demonstrates a profound advancement in planning, observation, and responsive actions, closely mirroring human cognitive processes \cite{touvron2023llama,wang2024survey, xi2023rise}.

% enabling a human-like approach to problem-solving. These systems, by breaking down complex tasks into manageable subtasks, demonstrate an advanced capability for planning, observation and responsive actions, thereby mirroring human cognitive processess more closely than ever before \cite{wang2024survey,xi2023rise,touvron2023llama}.
The LangChain Agent \cite{topsakal2023creating} further extends these capabilities by enabling LLMs to sequence actions, thus expanding the potential for automation. This strategic utilization of LLMs underscores a move towards more sophisticated, context-aware AI systems.

Developments in multi-agent systems, such as AutoGen \cite{wu2023autogen}, MetaGPT \cite{hong2023metagpt}, and AutoAgents \cite{chen2023autoagents}, introduce a novel dimension to the field by enabling collaboration and competition among agents. This architecture leverages the unique strengths of individual agents, enhancing system-wide efficiency and adaptability in complex scenarios. Such collaborative frameworks are proving essential for large-scale applications in sectors like healthcare \cite{dhatterwal2023multi,Jones2022Role}, finance \cite{gautier2023risk}, and urban planning \cite{bae2022hybrid}.

The versatility and efficacy of these LLM-based agents are highlighted in a range of application domains: In robotics, agents assist in both navigational tasks and complex manipulations, contributing to more autonomous and adaptive robotic systems \cite{driess2023palme}. Web manipulation and data extraction have seen significant enhancements, enabling better data-driven decision-making and user interaction models \cite{yao2022webshop, zhou2023webarena}. The gaming industry benefits from more dynamic AI opponents and game management systems, which improve user engagement and game design \cite{fan2022minedojo, tan2024towards}. Automated data analysis tools powered by LLM agents offer sophisticated insights and predictions, facilitating more effective data handling in business and research \cite{wang2023voyager, zhang2023data}.

\subsection{Autonomous agents}

The application of LVM systems for interfacing with Graphical User Interfaces (GUIs) in digital applications have been a key area of research. Wonderland \cite{yan2023gpt} harnessed the capabilities of GPT-4V which delineated by Yang et al. \cite{yang2023dawn} to navigate through mobile application interfaces by analyzing screenshots. Concurrently, the AppAgent \cite{yang2023appagent} utilizes GPT-4V to simulate the actions of mobile users, facilitation the execution of tasks in mobile applications through the analysis of device snapshots. Moreover, the MobileAgent incorporates Optical Character Recognition (OCR) technology to enhance the capabilities of of GPT-4V within a similar mobile agent architecture. This strategic enhancement allows MobileAgent to attain task completion rates on par with human operators. In contrast, CogAgent \cite{hong2023cogagent} constructs a specifically engineered LVM for interpreting and navigating GUIs. UFO \cite{zhang2024ufo} utilizes the windows inspect tool to acquire component and control information from system applications, providing it as context to GPT-4V to fully leverage its planning capabilities. 

\section{MMAC-Copilot Framework}

The workflow of MMAC-Copilot is designed to leverage the specialized capabilities of each agent in a sequential and collaborative manner. Here, we present an algorithmic overview in Algorithm \ref{algori:overview}. In section 3.1, we discuss the specialization of agents within the framework, each equipped with distinct roles and tools. In section 3.2, we delve into communications protocols which bolster accuracy in agents collaborations. In section 3.3, we introduce the team collaboration chain, dynamic framework that utilizes continuous feedback and iterative improvements to adapt framework strategies in real-time.

\begin{algorithm}[t]
\caption{Workflow of MMAC-Copilot}\label{algori:overview}
\begin{algorithmic}[1]

\Require User request $R$
\Ensure Task completion that satisfies $R$
\State Initialize system state $S_0$
\State Planner formulates initial plan $P_0$ based on $R$
\For{each subtask $s$ in $P_0$}
    \State Assign $s$ to appropriate agent $A$
    \State $A$ executes $s$, updates system state to $S_i$
    \State Mentor checks $S_i$; provides feedback $F_i$
    \If{$F_i$ indicates adjustment needed}
        \State Planner revises $P_i$ based on $F_i$
        \State Go to step 4 with revised $P_i$
    \ElsIf {$F_i$ indicates no adjustments are needed}
        \If{$s$ is the last subtask in $P_0$}
            \State Terminate loop
        \Else
            \State Continue to the next subtask
        \EndIf
    \EndIf
\EndFor

\end{algorithmic}
\end{algorithm}

% \begin{algorithm}
% \caption{Workflow of MMAC-Copilot}
% \begin{algorithmic}[1]
% \Require User request $R$
% \Ensure Task completion that satisfies $R$
% \State Initialize system state $S_0$
% \State Planner formulates initial plan $P_0$ based on $R$
% \For{each subtask $s$ in $P_0$}
%     \State Assign $s$ to appropriate agent $A$
%     \Repeat
%         \State $A$ executes $s$, updates system state to $S_i$
%         \State Mentor checks $S_i$; provides feedback $F_i$
%         \If{$F_i$ indicates adjustment needed}
%             \State Planner revises $P_i$ based on $F_i$
%             \State Go to step 5 with revised $P_i$
%         \EndIf
%         \State Verify if $S_i$ satisfies $R$
%     \Until{Satisfied}
%     \State Break loop and finalize task
% \EndFor
% \end{algorithmic}
% \end{algorithm}

\subsection{Specialization of agents}
Empirical studies suggest that diversity within human groups contributes to a variety of perspectives, which in turn enhances the group's performance across various tasks \cite{bransford1993ideal,williams1998demography,woolley2015collective}.
Also, contemporary research indicates that assigning specific roles to autonomous agents can improve their effectiveness \cite{li2023camel,qian2023communicative,salewski2024context}.
 Thus, in our framework, we have defined six agents, namely: Planner, Librarian, Programmer, Viewer, Video Analyst, and Mentor, as shown in Figure \ref{fig:framework}. The followings are detailed introductions to each agent:
%Meanwhile, we have specified the profiles for these agents, which include their names, memory, targets and expertise. Additionally, we intialize the agents with the different tools and capability. for example, "Programmer" possesses the capability to code and execute program (python and bash command).  "Viewer" have the tools to understand the screenshot and click the UI component.

 \textbf{Planner}: Employs LLM to strategically manage and allocate tasks among agents, optimizing workflow efficiency. In contrast to other frameworks, our Planner is not required to decompose tasks into indivisible atomic subtasks. We recognize Planner operating solely in the textual modality may not effectively integrate visual information such as UI elements, potentially leading to hallucination. Therefore, when Planner identifies a subtask which necessitates the incorporation of visual information, it provides only a coarse outline of the subtask. It allows each agent to leverage its own expertise, enhancing the adaptability of the overall framework.
   
\textbf{Librarian}: Utilizes LLM and API for information retrieval, enabling it to answer queries and provide foundational knowledge. 
    %It can process vast amounts of information quickly, supporting user requests and provide full context with other content.
It possesses the capability to rapidly process vast amounts of information, thereby supporting user requests and providing related context with other agent.

\textbf{Programmer}: Be responsible for writing and executing scripts, such as Python and Bash, to directly interact with software environments. This role is essential for tailoring applications to meet diverse requirements. The Programmer refines existing code by diagnosing and resolving issues. This process involves clarifying identified problems, enhancing code functionality, handling exceptions better, and providing detailed explanations.

Accurate Python statements are generated by understanding the context of Python classes and analyzing method parameters, ensuring syntactically correct calls. Also a evaluation mechanism further complements the Programmer's capabilities, assessing the refined code against task requirements, its functionality, and alignment, leading to a comprehensive review process. More detailed Prompt has been listed on Appendix \ref{sec:appendix:prompts}. 

The interactions within the Programmer's mechanism are mathematically represented as:
\[
C_{\text{init}} \xrightarrow[R]{E} C_{\text{mod}} \xrightarrow[X]{\delta} \text{Out} \xrightarrow[V]{T} (\text{Judge}, \text{Score})
\]
where:
\begin{itemize}
    \item $C_{\text{init}}$ is the initial code.
    \item $E$ stands for error analysis.
    \item $R$ is the refinement process.
    \item $C_{\text{mod}}$ represents the modified code.
    \item $\delta$ includes environmental variables.
    \item $X$ is the execution function.
    \item $\text{Out}$ includes runtime outputs and errors.
    \item $T$ corresponds to the original task requirements.
    \item $V$ is the evaluation process.
    \item $\text{Judge}$ determines if the task is accomplished.
    \item $\text{Score}$ assesses the code's generality and robustness, with higher scores indicating retention.
\end{itemize}

\textbf{Viewer}: The Viewer module integrates GPT-4V for the sophisticated interpretation of complex visual data extracted from screenshots. By coupling this capability with the SeeClick model \cite{cheng2024seeclick}, the Viewer effectively translates visual analyses into executable commands—such as initiating clicks on identified user interface elements. This integration bridges the gap between visual comprehension and application interaction, enabling automated manipulation of graphical interfaces based on visual understanding.

\textbf{Video Analyst}: The Video Analyst leverages the Gemini Vision to process and analyze video content, extracting critical visual information relevant to the task domain. This module is essential for subtasks that depend on video-based data, enhancing the model's ability to generalize across different modalities. By interpreting queries and retrieving pertinent details from video streams, the Video Analyst delivers precise and contextually appropriate responses. These capabilities are crucial for providing comprehensive contextual insights and assisting other agents in understanding the current operational state. Consequently, the Video Analyst adeptly handles complex scenarios where video data is paramount.

\textbf{Mentor}: Powered by GPT-4V, the Mentor agent plays a pivotal role in overseeing the system's execution processes, providing strategic oversight and troubleshooting support. After each subtask is executed, the Mentor analyzes screenshots from the current application window, offering a detailed description of the visible state. It evaluates whether the intended changes or commands have successfully impacted the application as expected. If the analysis indicates that adjustments are needed, the Mentor generates feedback based on these observations, which is then incorporated into the system’s planning process for the next iteration.

This feedback loop allows the planner to revise the current plan and reattempt the adjusted task. However, if no adjustments are necessary, the system proceeds to the next subtask in the sequence. In the case where the last subtask is reached and no further changes are required, the system terminates the process. This iterative mechanism of feedback and improvement not only ensures the successful completion of tasks but also enhances the overall system performance by continuously refining actions based on empirical data.

%展示其 //
\subsection{Communication Protocols}
Following established multi-agent frameworks \cite{hong2023metagpt,li2023camel,park2023generative,zhang2023building,zhuge2023mindstorms}, MMAC-Copilot adopts Structured Communication Interfaces, avoiding the traps of unconstrained nature language communication. Each agent communicates through a defined schema appropriate for their role. 
%, preventing the distortion of data typically seen in sequential natural language communications.  

In specific, we define the output format of the agents as JSON, and stipulate the keys which should be output according to the different roles of each agent. Meanwhile, each agent only receives the content of the required key. For example, "observation" key is exclusive output by Mentor and Viewer, as it is specifically to their role in emphasizing visual observations of screenshots. Apart from themselves, only the Planner receives the content of "observation" key, both when revising initial plans. 
%This kind of communication protocols ensure precise and efficient information exchange among agents, minimizing the risks of miscommunication inherent in sequential natural language communication.
These protocols are designed to prevent common issues such as data ambiguity in complex multi-agent interactions. By standardizing the communication format and defining clear protocols for message acknowledgment, MMAC-Copilot minimizes the risks of miscommunication and ensures which tasks are executed reliably. Additionally, the use of JSON allows for easy integration of new agents as the system scales, supporting the framework's adaptability and expansion.

\begin{figure}[h]
\centering
\includegraphics[width=0.5\textwidth,height=0.7\textheight,keepaspectratio]{ 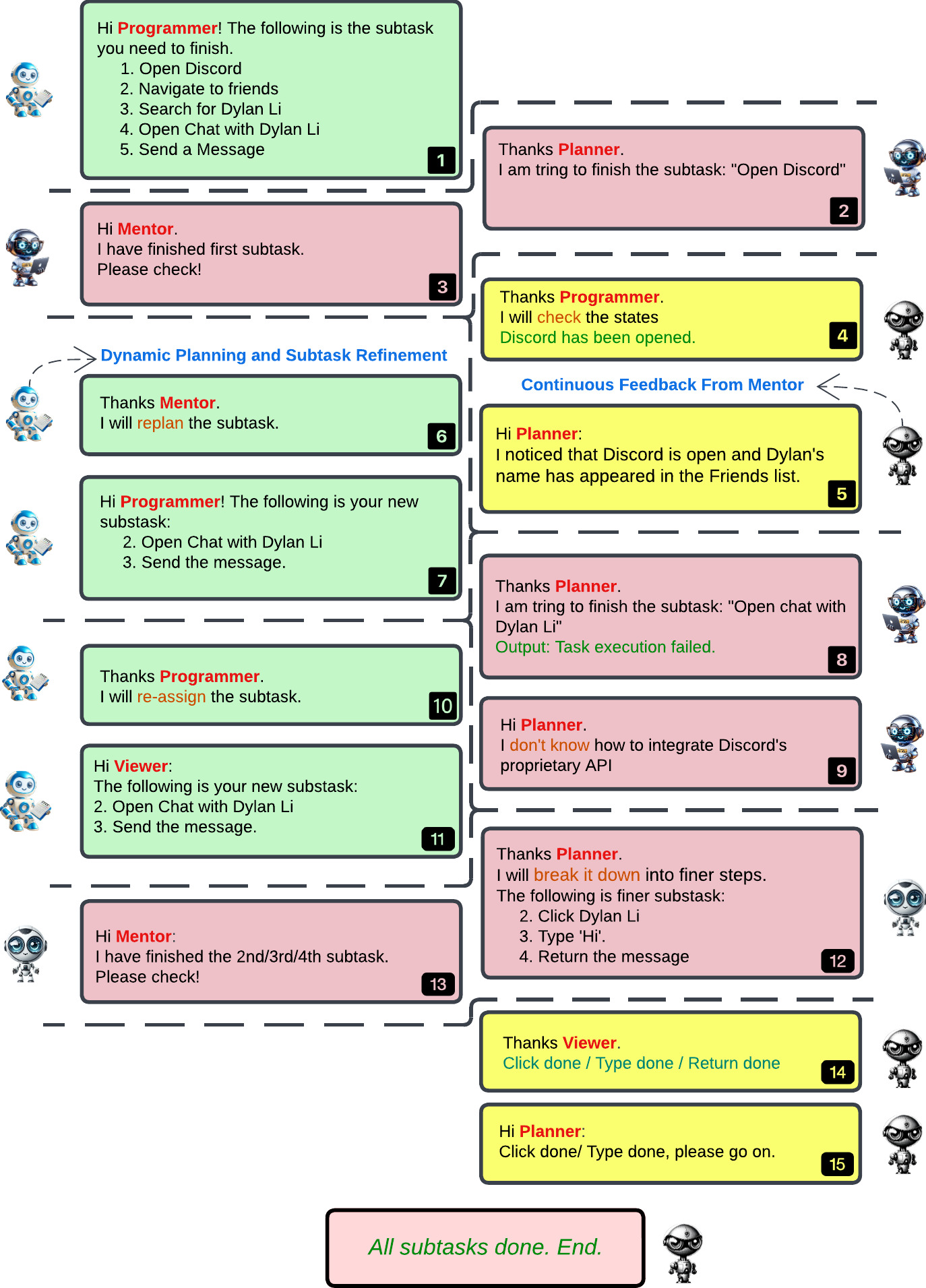}
\caption{Example of MMAC-Copilot. User query is "Open the Discord and Send a message to Dylan Li". \textcolor{red}{Red} represents the agent that will take the next action in the messaging process. \textcolor{brown}{Brown} indicates the current agent performing the task. \textcolor{green}{Green} color represents state feedback.}
\label{fig:dct}
\end{figure}

\subsection{Team Collaboration Chain}

Team collaboration chains leverage the strengths of dispersed management structures to facilitate enhanced cooperation among team members across different domains. By distributing responsibilities and decision-making power, these chains enable a more agile and responsive organizational environment. Such chain has been successfully implemented across multiple domains \cite{bang2005samarbejde, huh2019xr,selvanathan2018comparative,yatsuka2020collaboration}.

In the MMAC-Copilot framework, team collaboration chain is essential for managing  difficult tasks. This section delves into the collaborative mechanisms that enable dynamic planning and the iterative refinement of tasks, ensure effective execution and continuous improvement. A concrete demonstration of the collaboration chain in operation can be seen in Figure \ref{fig:dct}.

%The concept of "team collaboration chains" has been extensively implemented across multiple  

%Platforms like Wikipedia and GitHub have proven the efficacy of team models in managing complex, collaborative projects \cite{romero2015coordination}. Inspired by these successes, MMAC-Copilot  incorporates team collaboration chain designed to enhance agent cooperation without a central controlling entity.

\subsubsection{Dynamic Planning and Subtask Refinement}
During the planning phase,In the planning phase, the planner formulates an initial plan based on user requests.
Initial plan often outline the subtasks in a coarse manner, addressing the dynamic nature of visual environments and the inherent limitations of planning based on textual information alone. This initial planning stage sets the foundation for task execution but requires further refinement to address the specifics of the visual statement.

As the system transitions to the execution phase, agents equipped with specialized modalities, such as the Viewer and Video Analyst, assume pivotal roles. These agents reassess the current visual context and refine the coarsely defined subtasks into detailed, atomic subtasks. This step is crucial for adapting to real-time changes in the visual interface, ensuring that each task is executed with precision based on the latest visual information. This process showcases the strength of team collaboration, where different agents contribute uniquely to the refinement and execution of tasks.

\subsubsection{Continuous Feedback and Iterative Improvement}
Mentor plays a fundamental role in the team collaboration chain by establishing a continuous feedback loop with the Planner. After the execution of subtasks, Mentor evaluates the current state of the system through detailed analysis of screenshots and the visible state of the application. This evaluation determines whether the intended changes or commands have successfully impacted the application as expected.

If discrepancies or incomplete tasks are identified, Mentor generates insights and structured feedback based on a deep analysis of the outcomes. This feedback is crucial for revising the current plans and is incorporated into the next round of planning. 
% This feedback loop between the Mentor and the Planner not only ensures high-quality task completion but also facilitates adaptive learning, allowing the MMAC-Copilot system to evolve its strategies based on real-world interactions and outcomes. 

% \subsection{Workflow across agents}

\section{Experiments}
% \subsection{Experimental setting}
\begin{figure*}[h]
 \centering
    \includegraphics[width=\textwidth,height=0.4\textheight,keepaspectratio]{ 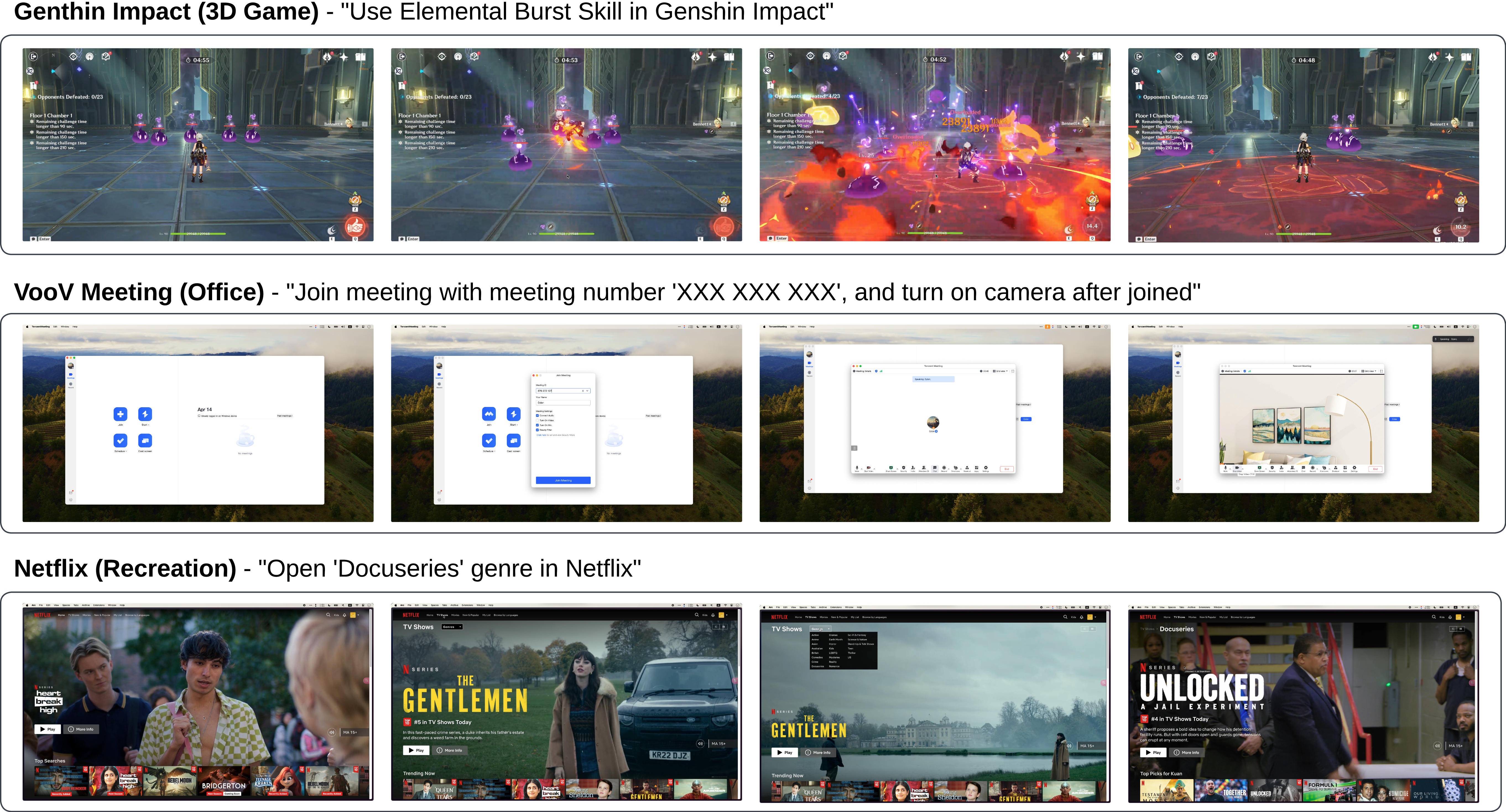}
    \caption{Task Completion under VIBench with MMAC-Copilot. Displayed is a sequential illustration of task completion across three distinct categories: (1) interactive gameplay performing skill in Genshin Impact; (2) operate within office application VooV Meeting; (3) recreational operation, exemplified by navigating in Netflix.}
    \label{fig:VIBench_example}
\end{figure*}
\subsection{Benchmark and setting}

\begin{table*}[h]
\caption{Performance comparison on GAIA. All performance are reported on the private test set.}\label{tab:GAIA}
\centering
\begin{tabular}{lc@{\hspace{4mm}}ccc} 
\toprule

\textbf{Model} & \textbf{Level 1} & \textbf{Level 2} & \textbf{Level 3} & \textbf{Average}  \\
\midrule
Human* & 93.90 & 91.80 & 87.30 & 91.00 \\
Search engine & 7.4 & 0.00 &0.00 & 2.47 \\
\midrule
% FRIDAY (Without learning) & 36.56 & 17.61 & 6.12 & 21.59 \\ 
GPT-3.5 \cite{achiam2023gpt} & 4.3 & 1.89 & 2.08 & 2.67 \\
GPT-4 \cite{achiam2023gpt}   & 9.68 & 1.89 & 0.00 & 4.00 \\
AutoGPT4 \cite{autogpt} & 15.05 & 0.63 & 0.00 & 5.00 \\
GPT4 Turbo \cite{achiam2023gpt}  & 9.68 & 6.92 & 0.00 & 6.67\\ 
GPT-4 Plugins \cite{achiam2023gpt} & 30.30 & 9.70 &0.00 &14.60\\
Chamomile \cite{Camomile} & 16.13 & 17.61 & 2.08 & 14.67 \\
FRIDAY \cite{wu2024oscopilot} & 40.86 & 20.13 & 6.12 & 24.25  \\

\midrule
\textbf{Ours} & \textbf{45.16} & \textbf{20.75} & \textbf{6.12} & \textbf{25.91~\stdvu{+6.8\%}} \\
\bottomrule
\end{tabular}
\end{table*}

\begin{table}[h]
\caption{Evaluation results on VIBench.}\label{tab:VIBench}
\centering
\resizebox{\columnwidth}{!}{
\begin{tabular}{lcccc}
\toprule
\textbf{Model} & \textbf{3D} & \textbf{Recreation} & \textbf{Office} & \textbf{Average}  \\
\midrule
UFO  & 0.00 & 28.57 & 15.38 & 14.65 \\
FRIDAY & 31.58 & 42.86 & 30.77  & 35.07  \\
\midrule
\textbf{Ours} & \textbf{63.16} & \textbf{69.23} & \textbf{78.57} & \textbf{70.32~\stdvu{+35.25}} \\
\bottomrule
\end{tabular}
}
\end{table}

We evaluate the performance of MMAC-Copilot using two distinct benchmarks designed to challenge and assess the capabilities of autonomous agent in diverse interaction environments.
The first is the General AI Assistant Benchmark (GAIA) \cite{mialon2023gaia}, which includes 466 rigorous question-answering tasks. GAIA primarily tests AI assistants in a question-answering (QA) format, focusing on their ability to utilize API calls for information retrieval and task execution. However, GAIA restricts the evaluation to textual interactions and API-based actions, which do not fully capture the broader range of GUI interactions that are prevalent in real-world applications.  

To address these limitations and to evaluate the agents' capabilities in a more complex and visually oriented environment, we introduced the Visual Interaction Benchmark (\textbf{VIBench}). VIBench comprises cases from three categories of applications: 3D gaming, recreation, and office environments. Unlike GAIA, VIBench emphasizes the necessity for agents to interact with applications through graphical user interfaces (GUIs) that cannot typically be controlled via standard APIs such as Win32 APIs.
The core interaction process in VIBench can be mathematically described as follows: Let \( S_0 \) represent the initial system state which includes the current GUI appearance and available actions. Given a user request \( R \), the system must transition through a series of states \( S_1, S_2, \ldots, S_n \) via actions \( a_1, a_2, \ldots, a_n \), ultimately achieving the final state \( S_f \) that satisfies \( R \). This process is represented by the sequence:
\[
S_0 \xrightarrow{a_1} S_1 \xrightarrow{a_2} S_2 \xrightarrow{\cdots} S_n \xrightarrow{a_n} S_f
\]
The primary goal of VIBench is not merely to evaluate the agent's ability to execute each step efficiently but to assess whether the final state $S_f$ satisfies the initial user request $R$. This objective emphasizes the effectiveness of the final outcome over the process, reflecting the practical focus of real-world applications:
\[
\text{Objective}: \text{Verify that } S_f \text{ meets } R
\]
% The primary goal of VIBench is to evaluate the agent's ability to execute the sequence $A$ effectively, leading to $S_f$ that satisfies the initial user request $R$. This objective can be represented as:
% \[
% \text{Objective}: \min \|A\| \text{ such that } S_0 \xrightarrow{A} S_f
% \]
% Here, $\|A\|$ denotes the complexity or the number of actions in sequence $A$, aiming to minimize this complexity while achieving the desired final state $S_f$. VIBench challenges the agent's planning, execution, and contextual understanding capabilities in a scenario that simulates typical user interactions with digital environments.

VIBench thus complements GAIA by adding a layer of complexity and realism, challenging agents to not only respond correctly in a QA format but also to effectively navigate and interact within a application environments that mirror everyday user challenges. Examples are shown in Figure \ref{fig:VIBench_example}. We also present the detailed list of request utilized in VIBench in Appendix \ref{sec:appendix:VIBench}.

\paragraph{Setting} To ensure a fair comparison with previous work, we employed the same versions of the ChatGPT API used in prior studies. Specifically, we utilized GPT-4-turbo-1106 and gpt-4-vision-preview. For the Video Analyst, we used gemini-1.5-pro-vision. More detailed experiment setting could been seen in Appendix \ref{sec:appendix:GAIA}.

\subsection{Evaluation Method}
For GAIA, each question calls for a specific response: string, number, or floats. Evaluation is conducted through a exact match between the model's answer and ground truth.  

For our VIBench, we adopt human evaluation which follow the SIMA evaluation method \cite{sima2024scaling}, recognizing that in numerous instances, task success cannot be automatically inferred. We engage human judges who possess expert-level familiarity with the specific applications, defined as having at least 20 hours of application usage. These experts are asked to assess recorded videos to determine whether the user requests have been accurately executed. 
During the assessment, human experts are instructed to disregard any irrelevant actions performed by the agents. Their judgments are solely based on whether the stated objectives of the tasks are achieved, without considering the intermediate steps or unrelated actions. Meanwhile, to minimize human subjectivity, we have imposed a maximum limit of 30 execution rounds for task. If the task is not completed within 30 rounds, it is judged as a failure.

% We use human judges who are application experts, i.e., they have used the specific application at least 20 hours. We ask them to review recorded videos to judge wheter the user request has been done.
\subsection{Baseline}
For GAIA, we present the performance of GPT-3.5 and GPT-4 with and without manually set plugins, along with results from AutoGPT, Chamonmile and FRIDAY which all use GPT-4 as their backend. The selection of Plugins for GPT-4 is conducted manually, with individual choosing based on specific user query. For comparative purposes, we also include data on human performance sourced from GAIA \cite{mialon2023gaia}.  
% rely on humans to browese and select proper plugins based on the task question.

For VIBench, we report the performance of UFO and FRIDAY. Notably, since FRIDAY does not natively support the Windows platform, we have modified FRIDAY's code to adapt it to Windows. Also we set UFO's Vision model same as our framework.
\subsection{Results}

MMAC-Copilot shows significant performance improvement across all levels of GAIA. As illustrated in Table \ref{tab:GAIA}, MMAC-Copilot achieved the highest scores in Level 1 (45.16) and Level 2 (20.75) tasks, matching the highest score in Level 3 tasks (6.12). Also on average, it outperformed the closest competing system, FRIDAY, by 6.8\% in  overall task performance. Given that tasks in GAIA encompass various complex AI assistant capabilities, from basic information retrieval to interactive task requiring multiple steps and integration of external data sources.

In VIBench as shown in Table \ref{tab:VIBench}, MMAC-Copilot outperformed other systems across all application categories (3D gaming, recreation, and office scenarios). Our model achieved average scores of 63.16\% in 3D gaming, 69.23\% in recreation, and 78.57\% in office applications, with an overall average of 70.32\%. It indicates MMAC-Copilot's robust capability in navigating and interacting within complex GUI environments, where direct API interaction is not possible.

% \textbf{Qualitative Analysis}

\section{Conclusion}
MMAC-Copilot is a novel framework enhances agent interactions within complex digital environments. This framework effectively addresses agent hallucination, ensuring reliable responses to environmental changes. Tested on the GAIA benchmark and our VIBench, MMAC-Copilot demonstrates superior performance compared to existing systems.

% \section*{Acknowledgments}
\section*{Limitation}
We acknowledge that while MMAC-Copilot has achieved a level of generality in application through the integration of GUI models for visual detection, it is still limited by the GUI model's ability to understand UI components. Consequently, MMAC-Copilot struggles with interpreting complex UI interfaces and performing multi-step operations that depend on such complexity. 
Additionally, MMAC-Copilot's capabilities in real-time 3D gaming environments reveal gaps in the spatial understanding, which is crucial for tasks such as navigation, aiming and shooting in 3D environments. The framework’s longer inference time impair the ability to perform tasks with higher temporal sensitivity.

To solve these, we suggest integrating external knowledge databases via online search engines, providing additional context for decision-making processes. This approach will not only help in understanding complex UI layouts but also in recognizing user intentions and environmental cues in real-time games.

\bibliography{custom}

\newpage

\begin{figure*}[h]
 \centering
    \includegraphics[width=\textwidth,height=0.4\textheight]{ 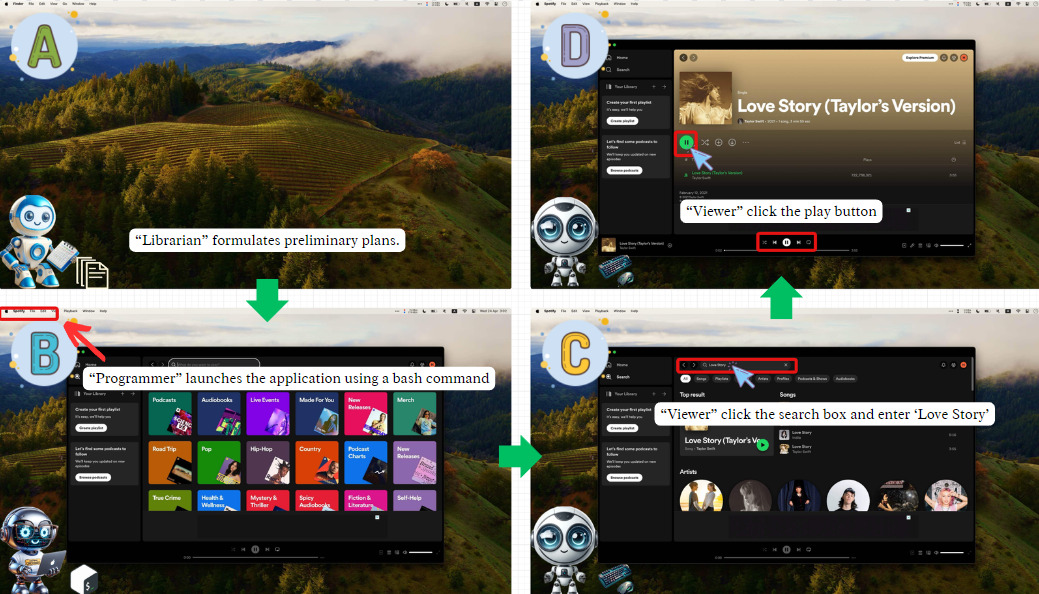}
    \caption{Illustrates the sequence of actions undertaken by MMAC-Copilot in response to the user's request "Play <Love Story> on Spotify". (A) Represents the initial state where the "Planner" formulates preliminary plans. . (B) "Programmer" launches the application using a bash command. (C) "Viewer" click the search box and enter <Love Story>. (D) "Viewer" click the play button.}
    \label{fig:teaser}
\end{figure*}

\appendix

\section{Demo on Mac OS}
The following Figure \ref{fig:teaser} illustrates an example of MMAC-Copilot performing an action sequence on macOS in response to the user request “Play <Love Story> on Spotify.” Each subfigure represents a distinct step in the process: (A) shows the “Planner” formulating the initial plan; (B) the “Programmer” launching the application using a bash command; (C) the “Viewer” clicking the search box and entering <Love Story>; and (D) the “Viewer” clicking the play button. This demonstrates how MMAC-Copilot efficiently handles a media playback request from the user.

\label{sec:appendix:demo}
\section{Agent prompts}
\label{sec:appendix:prompts}
\begin{figure*}[t]
    \centering
    \includegraphics[width=\textwidth]{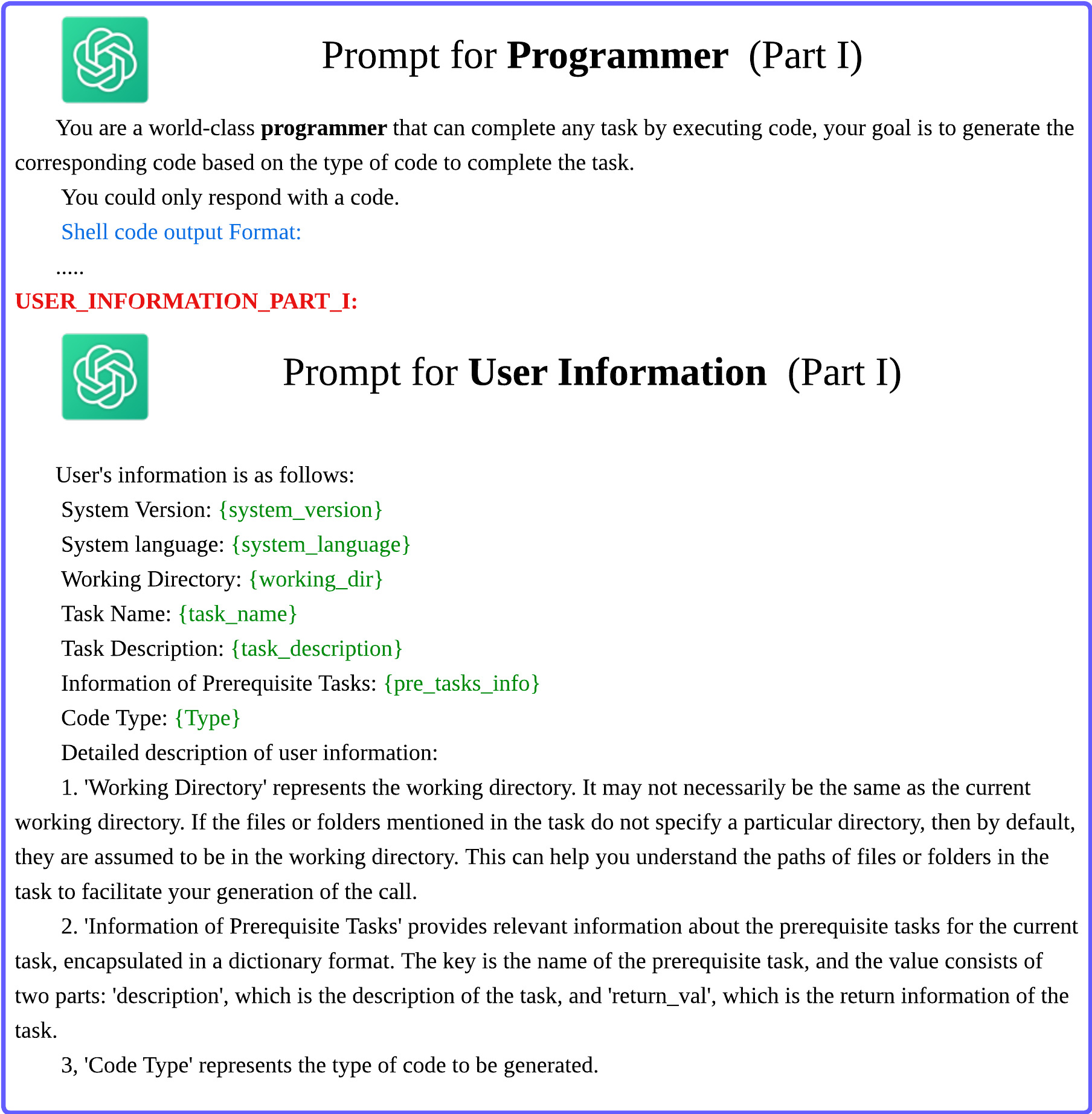}
    \caption{Prompt for Programmer Part}
    \label{fig:Programmer-prompt}
\end{figure*}

\begin{figure*}[t]
    \centering
    \includegraphics[width=\textwidth]{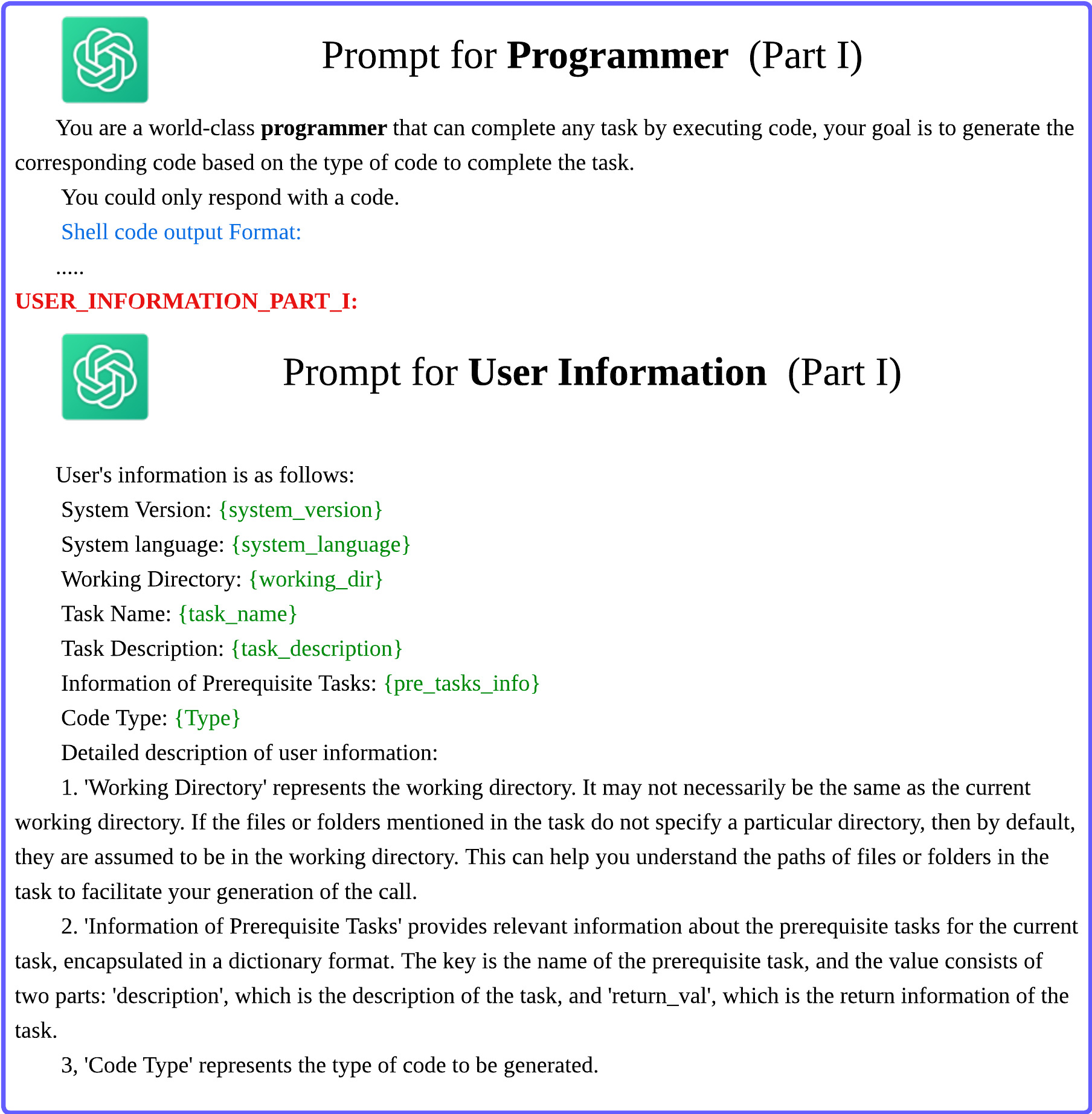}
    \caption{Prompt for Programmer Part}
    \label{fig:programmer-prompt-1}
\end{figure*}

\begin{figure*}[t]
    \centering
    \includegraphics[width=\textwidth]{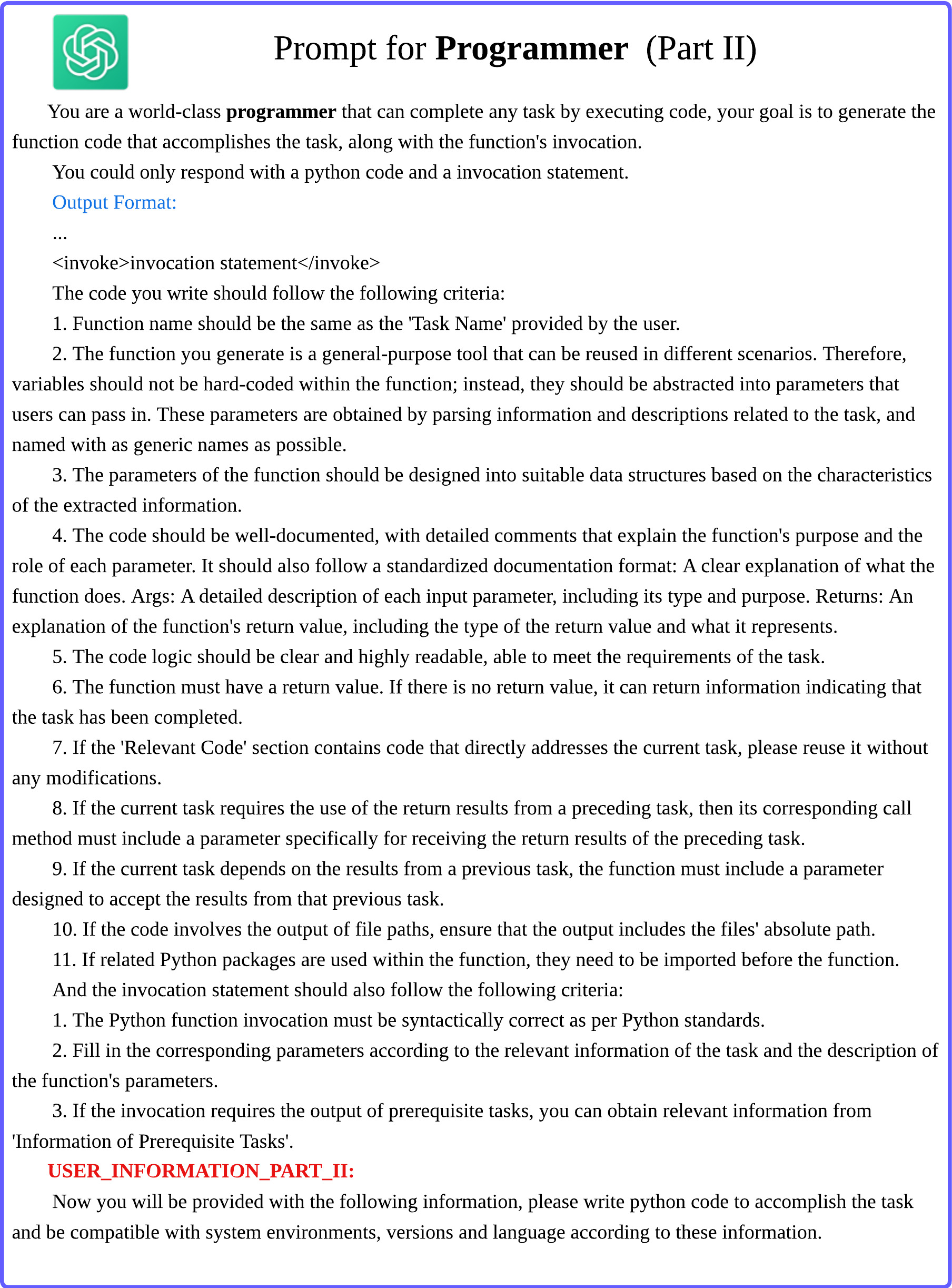}
    \caption{Prompt for Programmer Part}
    \label{fig:programeer-prompt-2}
\end{figure*}

\begin{figure*}[t]
    \centering
    \includegraphics[width=\textwidth]{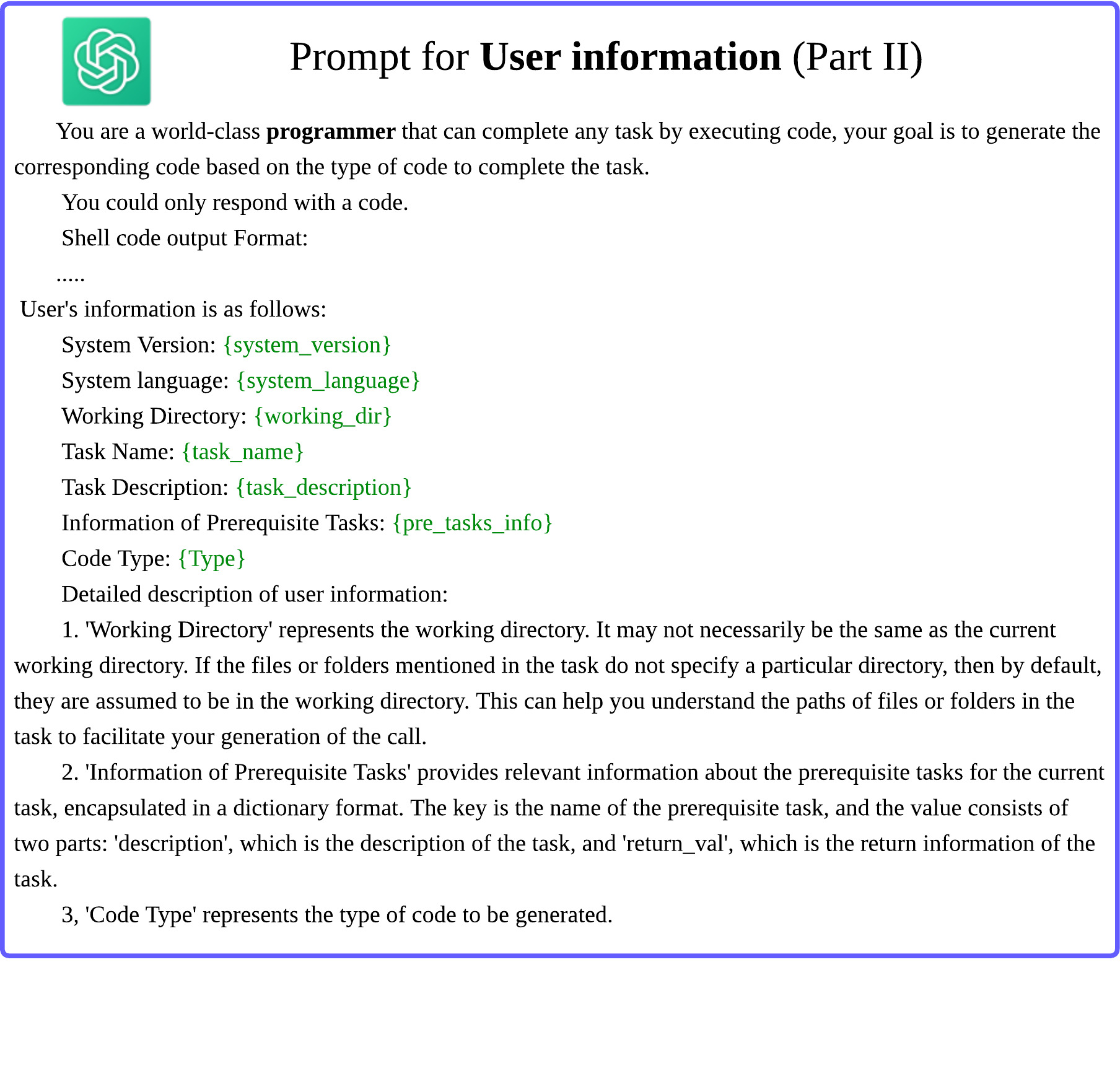}
    \caption{Prompt for Programmer Part}
    \label{fig:programmer-prompt-3}
\end{figure*}

\begin{figure*}[t]
    \centering
    \includegraphics[width=\textwidth]{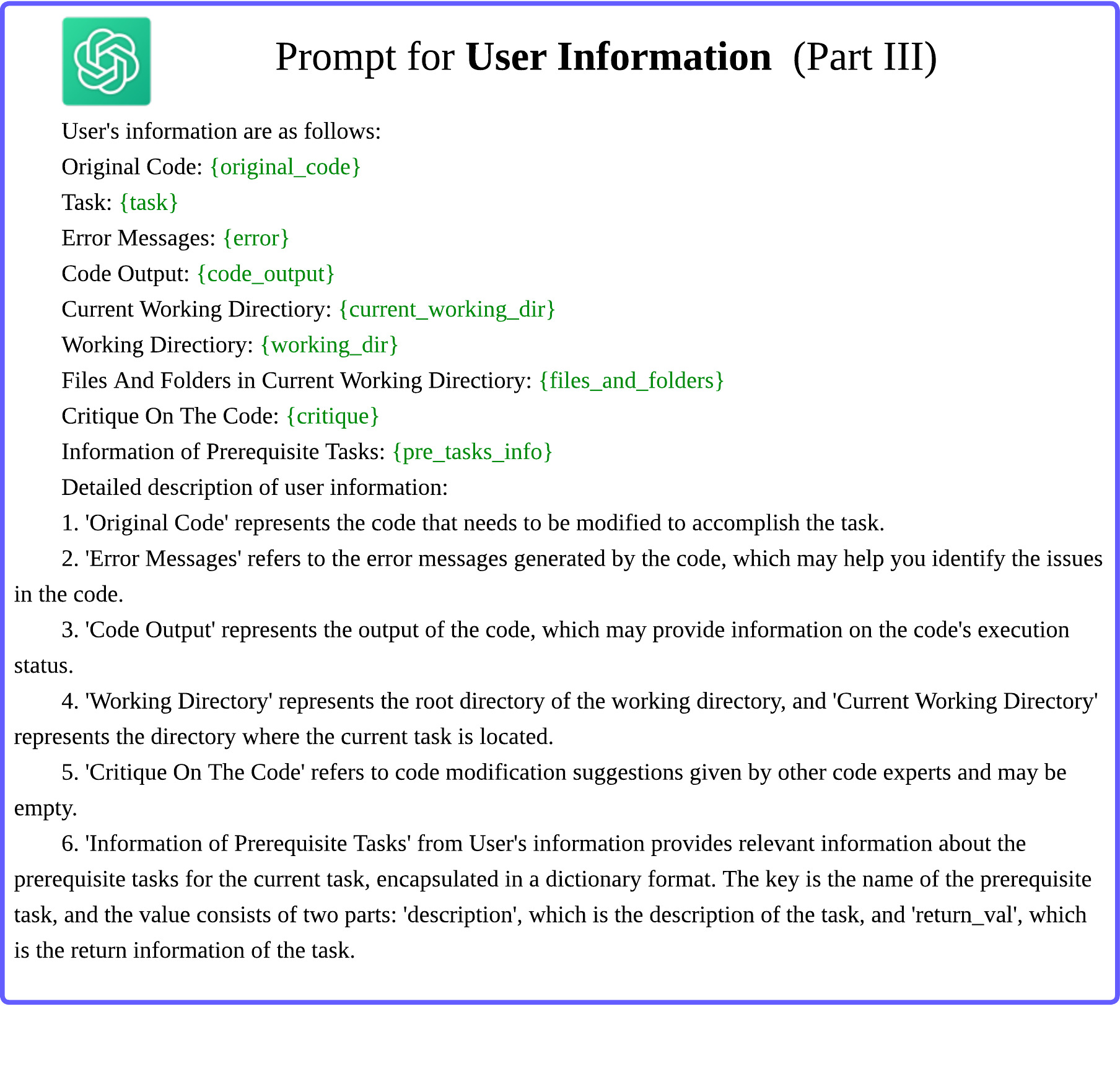}
    \caption{Prompt for Programmer Part}
    \label{fig:programmer-prompt-4}
\end{figure*}

\begin{figure*}[t]
    \centering
    \includegraphics[width=\textwidth]{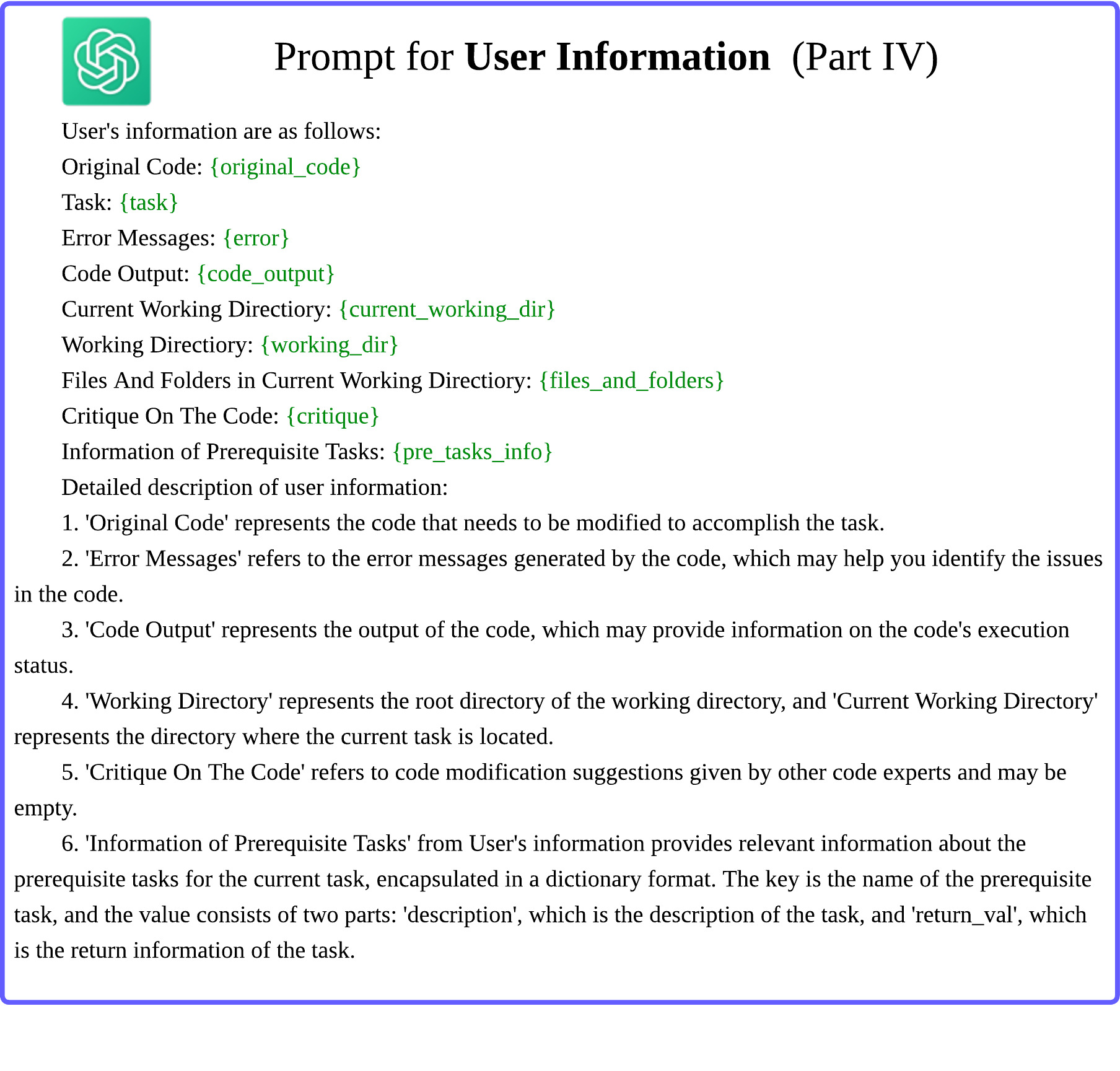}
    \caption{Prompt for Programmer Part}
    \label{fig:programmer-prompt-6}
\end{figure*}

\begin{figure*}[t]
    \centering
    \includegraphics[width=\textwidth,height=\textheight]{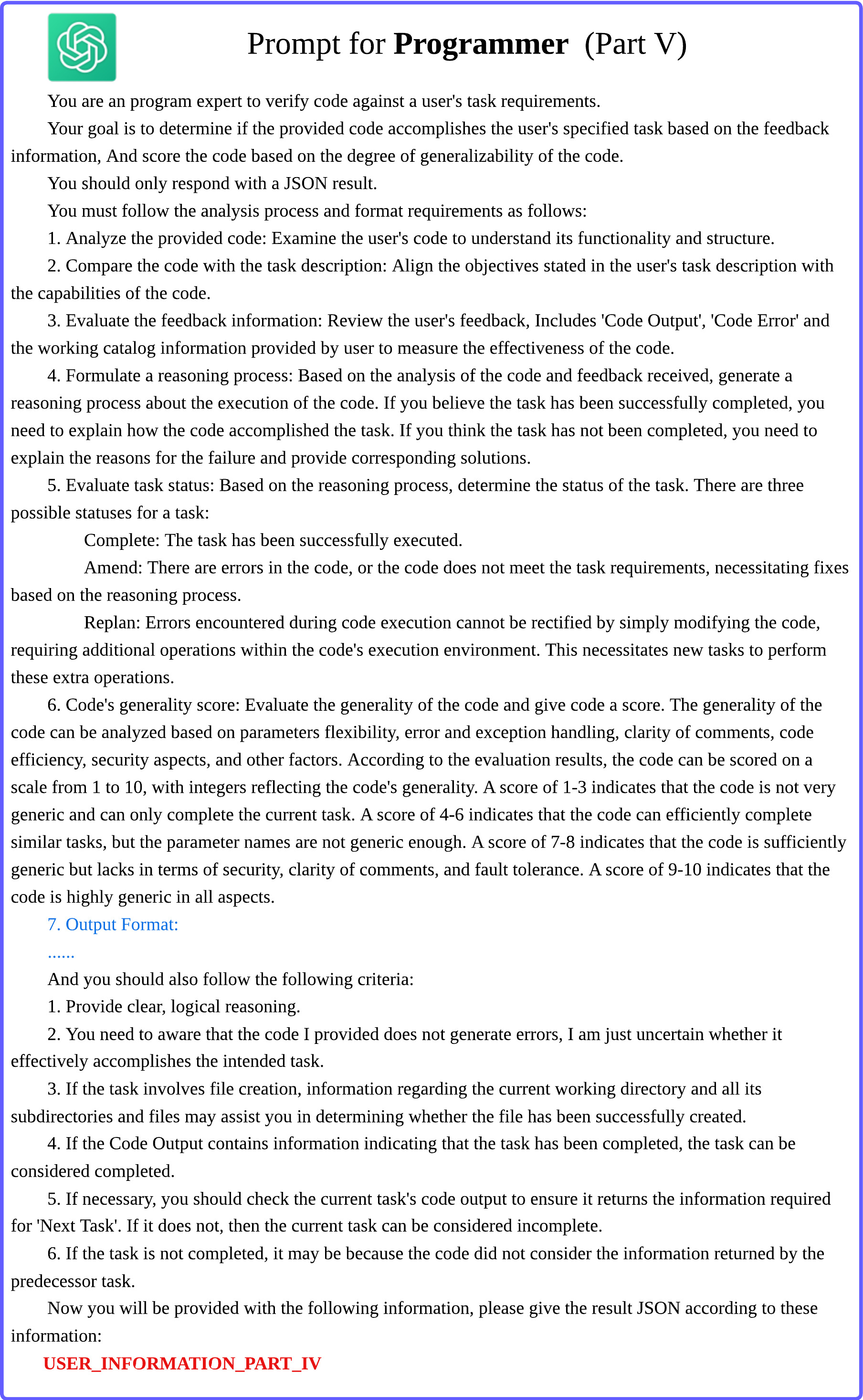}
    \caption{Prompt for Programmer Part}
    \label{fig:programmer-prompt-7}
\end{figure*}

\begin{figure*}[t]
    \centering
    \includegraphics[width=\textwidth]{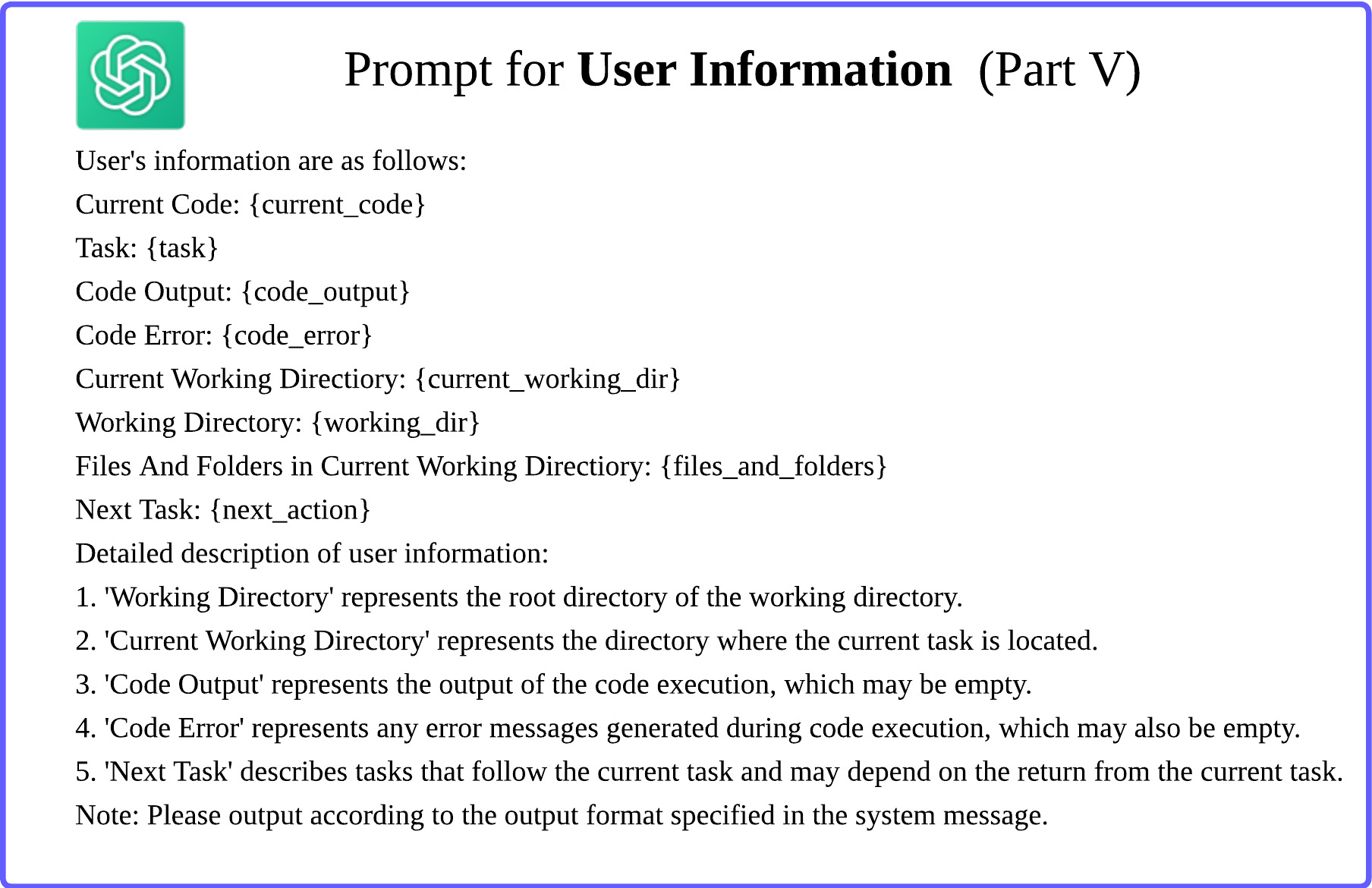}
    \caption{Prompt for Programmer Part}
    \label{fig:programmer-prompt-8}
\end{figure*}

To better explain the operating mechanism of Programmer, we have provided the complete prompts in the Figure \ref{fig:Programmer-prompt},\ref{fig:programmer-prompt-1},\ref{fig:programeer-prompt-2},\ref{fig:programmer-prompt-3},\ref{fig:programmer-prompt-4},\ref{fig:programmer-prompt-5},\ref{fig:programmer-prompt-6},\ref{fig:programmer-prompt-7} and \ref{fig:programmer-prompt-8}.

\section{GAIA Detail Experiment Setting}
\label{sec:appendix:GAIA}
GAIA serves as a comprehensive benchmark for evaluating general AI assistants by testing their ability to handle real-world scenarios across various capabilities, including advanced reasoning, multi-modality handling, coding, and web browsing. The benchmark comprises text-based questions, sometimes accompanied by supplementary files such as images or spreadsheets. These questions are designed to cover a broad range of use cases from everyday tasks to complex scientific queries, necessitating concise and unambiguously correct answers that are straightforward to verify.

\subsection{Levels of Difficulty}

The questions in GAIA are categorized into three levels based on their complexity:
\begin{itemize}
    \item \textbf{Level 1:} Simplest tasks, generally requiring minimal tool use and up to five steps.
    \item \textbf{Level 2:} More complex tasks that require integrating information from multiple sources, involving about 5 to 10 steps.
    \item \textbf{Level 3:} The most complex tasks, demanding extensive sequences of actions and the use of multiple tools to solve real-world problems.
\end{itemize}

\begin{figure*}[t]
 \centering
    \includegraphics[width=\textwidth]{ 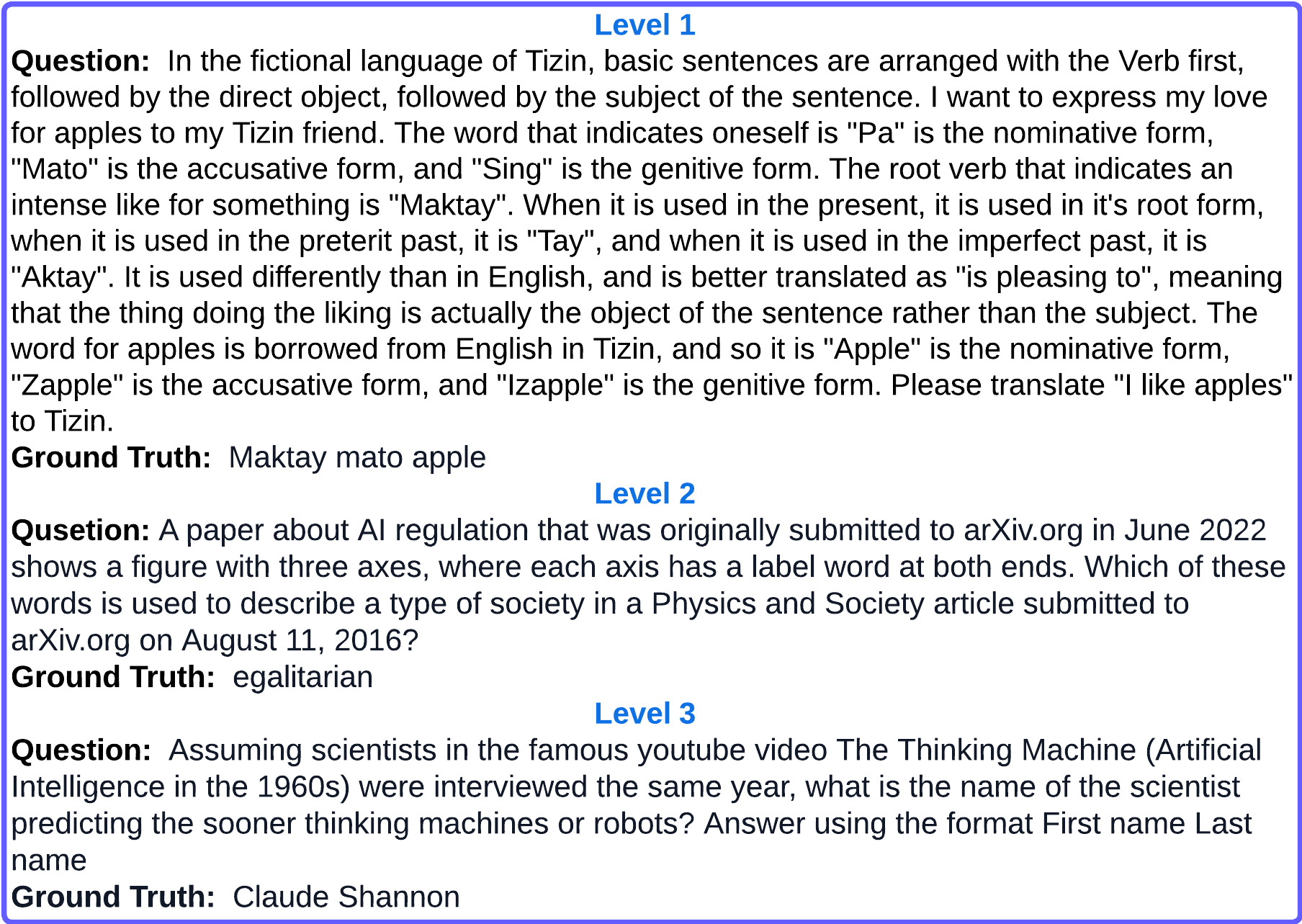}
    \caption{Sample Questions from GAIA. These tasks necessitate core competencies including reasoning, the ability to handle multiple modalities, and proficiency in using tools. The answers to these questions are unequivocally defined and are intentionally unlikely to be present in unstructured training datasets. Several questions are supplemented with visual evidence, mirroring authentic scenarios and enhancing the precision of query formulation.}
    \label{fig:GAIA-samples}
\end{figure*}

We present a sample GAIA question from Level 1 through Level 3 in Figure \ref{fig:GAIA-samples}

\subsection{Evaluation Methodology}

GAIA employs an automated, precise, and factual evaluation method. Each question demands an answer in a specified format (string, number, or comma-separated list) and correctness is verified through an exact match with a pre-defined ground truth, ensuring the evaluation's reliability and efficiency.

\subsection{Model Experimental Setup}
In this section, we introduce the detailed experiment setting of our comparison models.

We evaluates leading AI models, including GPT-3.5, GPT4 and AutoGPT, also conduct non-AI baselines experiments such as human annotators and traditional web search methods. We will introduce each experiment setting as following.

\begin{figure*}[t]
    \centering
    \includegraphics[width=\textwidth]{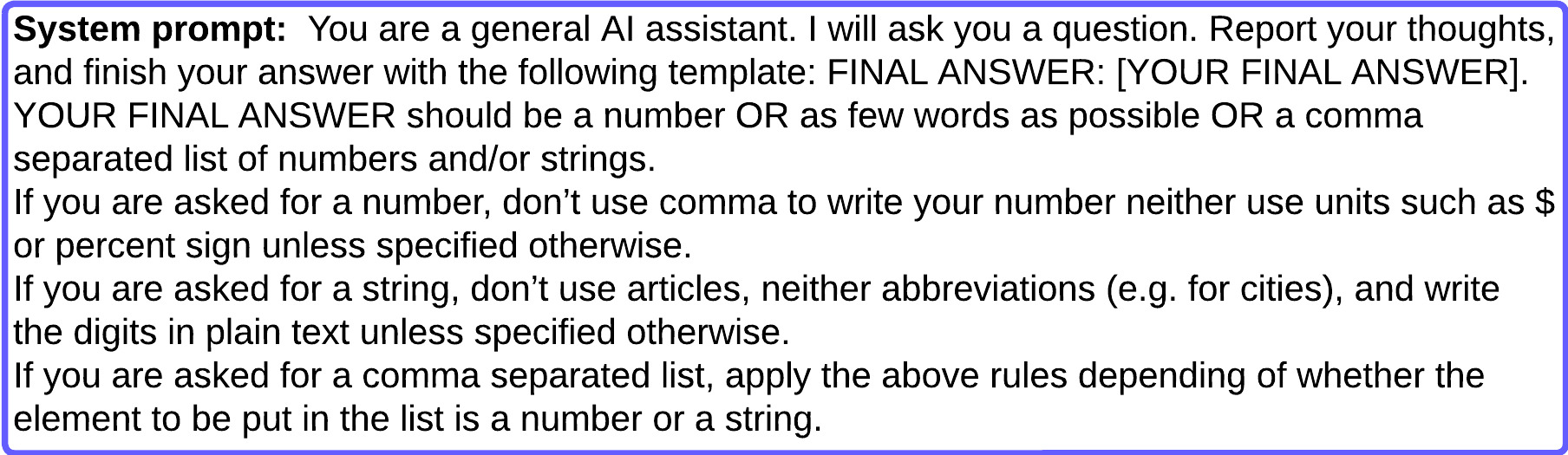}
    \caption{The system prompt to answer GAIA.}
    \label{fig:system-prompt}
\end{figure*}

\textbf{Comparison Models:}
\begin{description}
\item[ GPT3.5/GPT4:] This setup evaluates the core linguistic and reasoning capabilities of GPT4 without the aid of any external tools or plugins. We use a
prefix prompt before asking the model a question. To ease answer extraction, we specify a format in the prefix prompt which could been saw in Figure \ref{fig:system-prompt}.

\item[GPT4 with Plugins:] In this configuration, GPT4 is enhanced with selected plugins that extend its functionality. This includes tools for advanced data analysis, web browsing, and handling of third-party APIs. 
\item[AutoGPT:] Utilizing GPT4 as its backend, AutoGPT automatically selects plugins based on the context of the query. This model is tested for its efficiency in tool selection and its ability to autonomously adapt to complex query requirements.
\end{description}

\begin{description}
\item[Human*:] Consisting of experts in various fields, human annotators provide responses based on their expertise and research abilities. Their performance sets a high standard for AI models, especially in tasks requiring deep domain knowledge and nuanced understanding.
\item[Search engine:] Using conventional search engines, this baseline evaluates the effectiveness of straightforward information retrieval. Specifically, we enter our questions into a search engine and review the first page of results to see if the answer can be inferred.

\end{description}

\textbf{Notes:}
Each model undergoes rigorous testing across a series of structured tasks that range from simple to highly complex, categorized into three levels of difficulty as previously described. Models are evaluated based on accuracy, the relevance of the responses, and the efficiency of handling the queries. Each session is replicated three times to ensure consistency and reliability in the results.

\section{VIBench}

\subsection{VIBench Results List}
\label{sec:appendix:VIBench}
\definecolor{ao}{rgb}{0.0, 0.5, 0.0} % Define the 'ao' color (green)
\newcommand{\xmark}{\ding{55}}% % Define the cross symbol
\newcommand{\greencheck}{{\color{ao}\checkmark}} % Define the green checkmark
\newcommand{\redcross}{{\color{red}\xmark}} % Define the red cross

In Table~\ref{tab:req1}, we provide a comprehensive list of the user requests featured in VIBench, along with the corresponding performance of MMAC-Copilot. These requests cover a range of commonly used functions across eight popular applications. For clarity, requests that involve follow-up tasks are numbered sequentially. In the Success column, ``\greencheck'' denotes that MMAC-Copilot completes the request successfully, while ``\redcross'' indicates a failure to fufill the request. 

\begin{table*}[h]
\caption{The requests in VIBench and detailed results achieved by MMAC-Copilot}\label{tab:req1}
\centering
\begin{tabularx}{\textwidth}{>{\centering\arraybackslash}p{10cm}|>{\centering\arraybackslash}p{3.5cm}|c}
\hline
\textbf{Operation}&\textbf{Application}&\textbf{Success}\\ \hline\hline
Ride on vehicle [Press F]&PUBG&\greencheck\\ \hline
Probe shot [Press Left mouse]&PUBG&\greencheck\\ \hline
Picking up supplies [Identify the screen and press F]&PUBG&\greencheck\\ \hline
Assemble accessories and discard unnecessary ones&PUBG&\redcross\\ \hline\hline
Use E skill&Valorant&\greencheck\\ \hline
Shoot&Valorant&\greencheck\\ \hline
Climb up the rope&Valorant&\redcross\\ \hline
Absorb energy ball&Valorant&\redcross\\ \hline\hline
Take turns to release Q skills&Genshin&\redcross\\ \hline
Pick up the Divine Eye&Genshin&\greencheck\\ \hline
Freeze trap&Genshin&\redcross\\ \hline
Fishing&Genshin&\redcross\\ \hline
Climb&Genshin&\redcross\\ \hline\hline
Send message to xxx&DingTalk&\greencheck\\ \hline
Create meeting&DingTalk&\redcross\\ \hline
Create a group&DingTalk&\greencheck\\ \hline\hline
End meeting&TencentMeeting&\redcross\\ \hline
Turn on camera&TencentMeeting&\greencheck\\ \hline
Create a meeting&TencentMeeting&\greencheck\\ \hline
Join a meeting&TencentMeeting&\greencheck\\ \hline
Click the record bottom&TencentMeeting&\greencheck\\ \hline
Copy Invitation&TencentMeeting&\greencheck\\ \hline\hline
Open the library in steam&Steam&\greencheck\\ \hline
Open the friends interface&Steam&\greencheck\\ \hline
Click on store and then click on the search box&Steam&\redcross\\ \hline\hline
View current music and play it&Spotify&\greencheck\\ \hline
Search for a specific piece of music and save it&Spotify&\greencheck\\ \hline\hline
Open docuseries in Netflix&Netflix&\greencheck\\ \hline
Save XXX Series to my list&Netflix&\redcross\\ \hline
Remove XXX from my list&Netflix&\redcross\\ \hline
\end{tabularx}
\end{table*}

\subsection{Visualization}
\begin{figure*}[t]
    \centering
    \includegraphics[width=\textwidth]{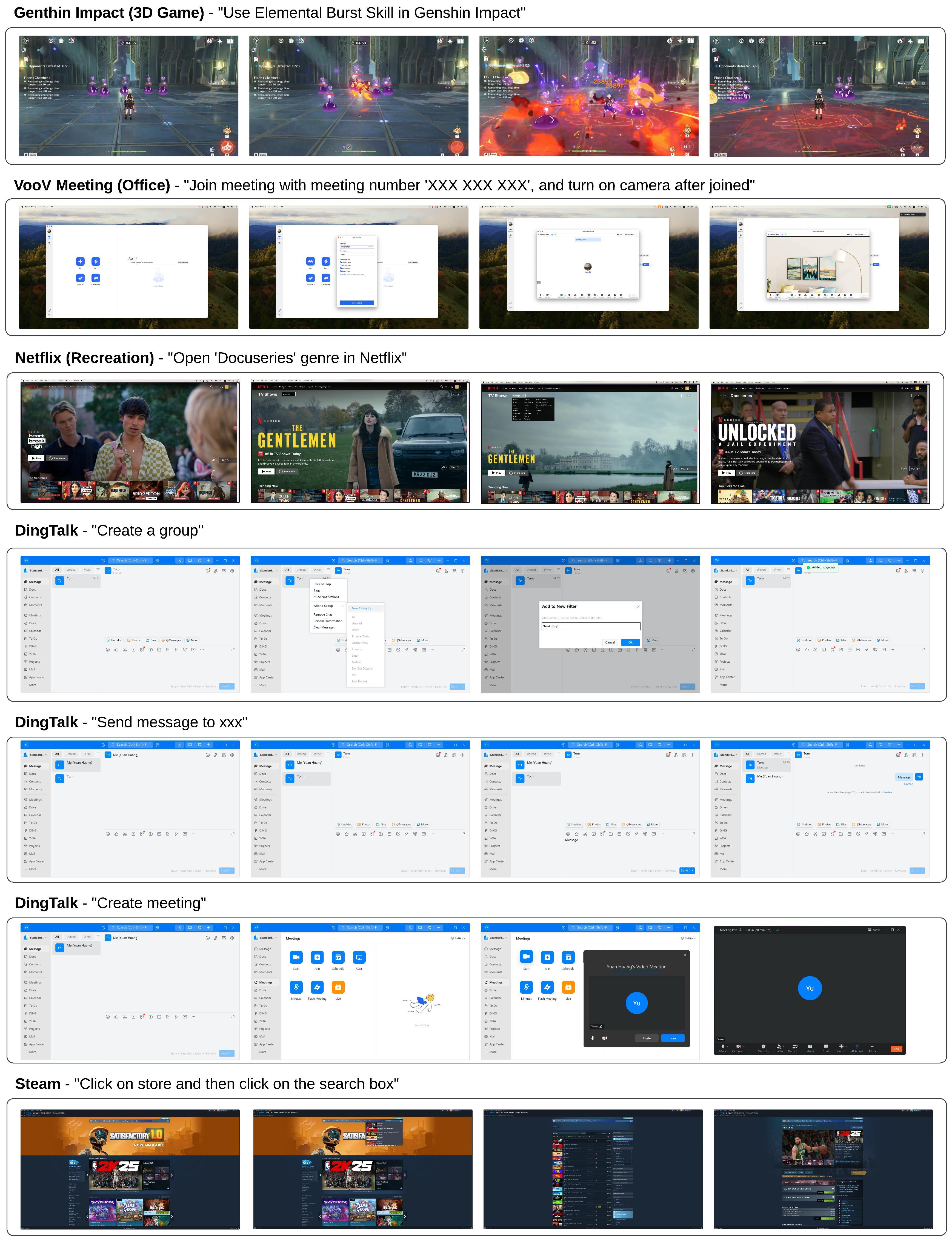}
    \caption{How VIBench works(Part-I)}
    \label{fig:VIBench-work-1}
\end{figure*}

\begin{figure*}[t]
    \centering
    \includegraphics[width=\textwidth]{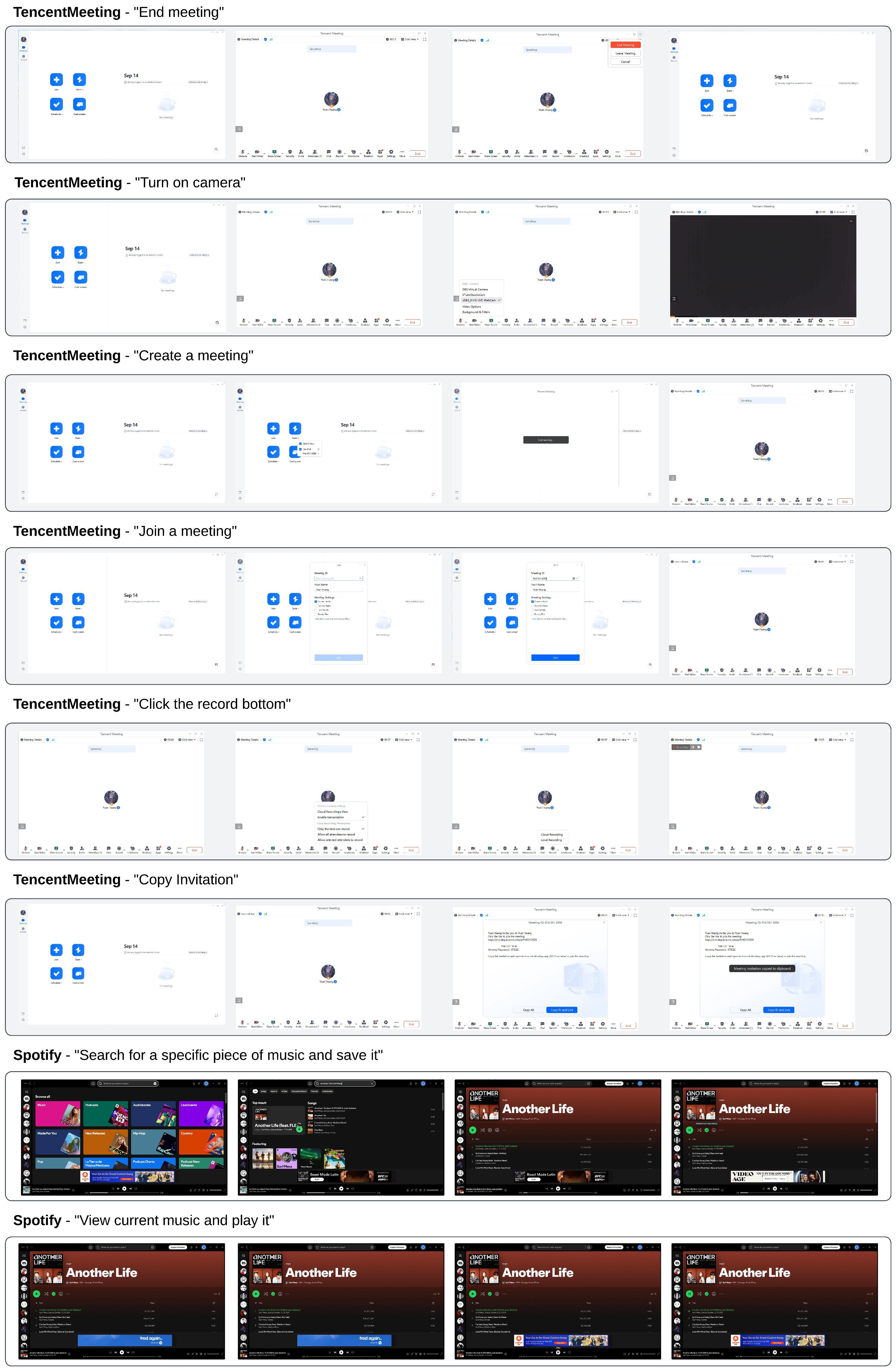}
    \caption{How VIBench works(Part-II)}
    \label{fig:VIBench-work-2}
\end{figure*}

We have visualized VIBench with requests in Figure \ref{fig:VIBench-work-1} and \ref{fig:VIBench-work-2}.

\subsection{Adapt for Windows}

To compatible with the Windows platform, we made some adaptations focusing on tool integration and modifications for platform-specific libraries. 

One of the primary adaptations involved the integration of tools such as the Python environment and Windows Shell support. The Python interpreter was configured to work within the Windows platform and Windows path handling were incorporated to ensure smooth execution of scripts and Python modules.
In addition to tool integration, some Python libraries required adaptation to function properly on Windows. We used alternative Python modules to replace database libraries that did not fully support the Windows environment for tool persistent. As Python libraries behave differently on Windows due to variations in concurrent requests management, we also made some adaptation for it.


\begin{figure*}[t]
    \centering
    \includegraphics[width=\textwidth]{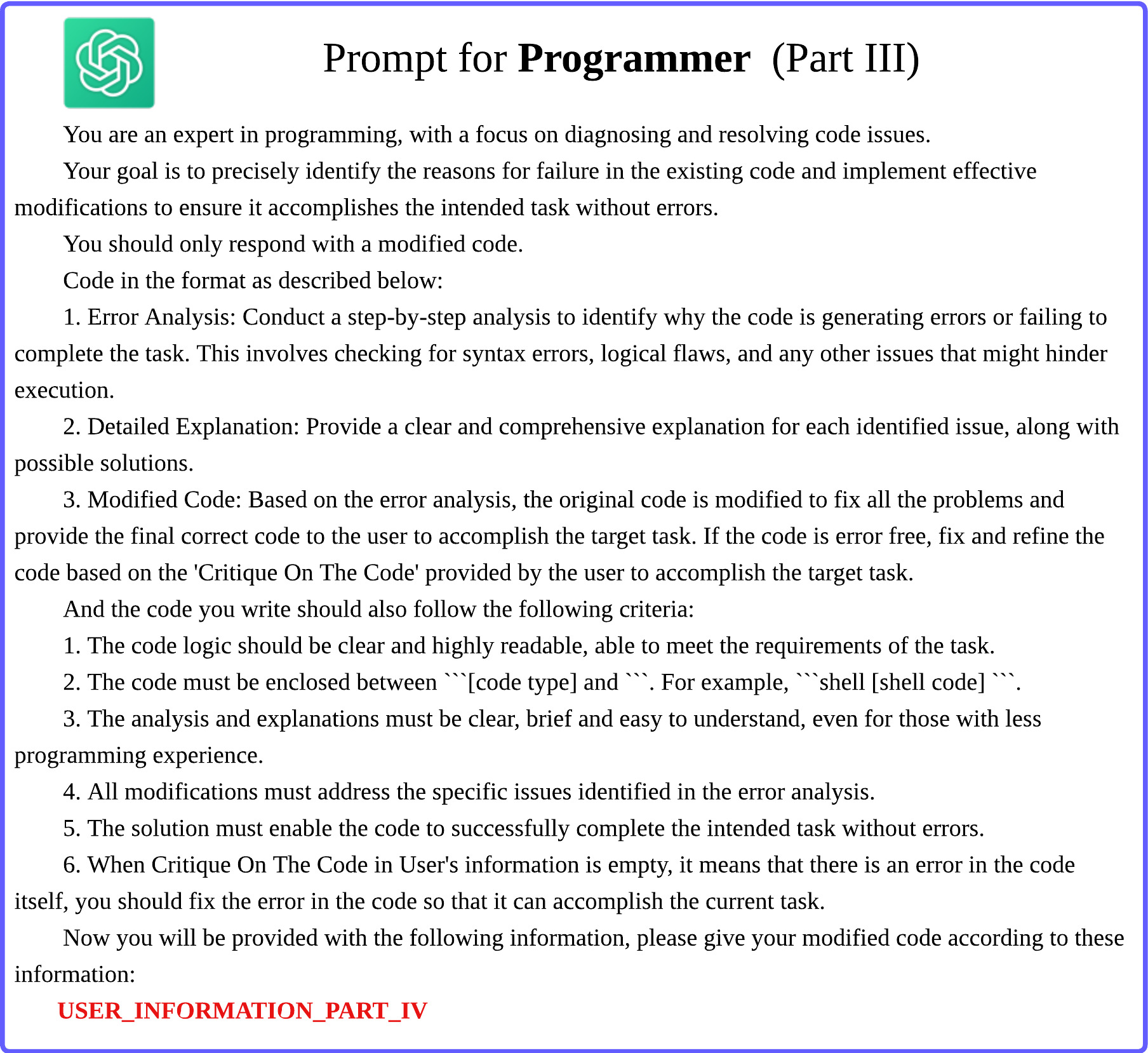}
    \caption{Prompt for Programmer Part}
    \label{fig:programmer-prompt-5}
\end{figure*}
\begin{figure*}[t]
    \centering
    \includegraphics[width=\textwidth]{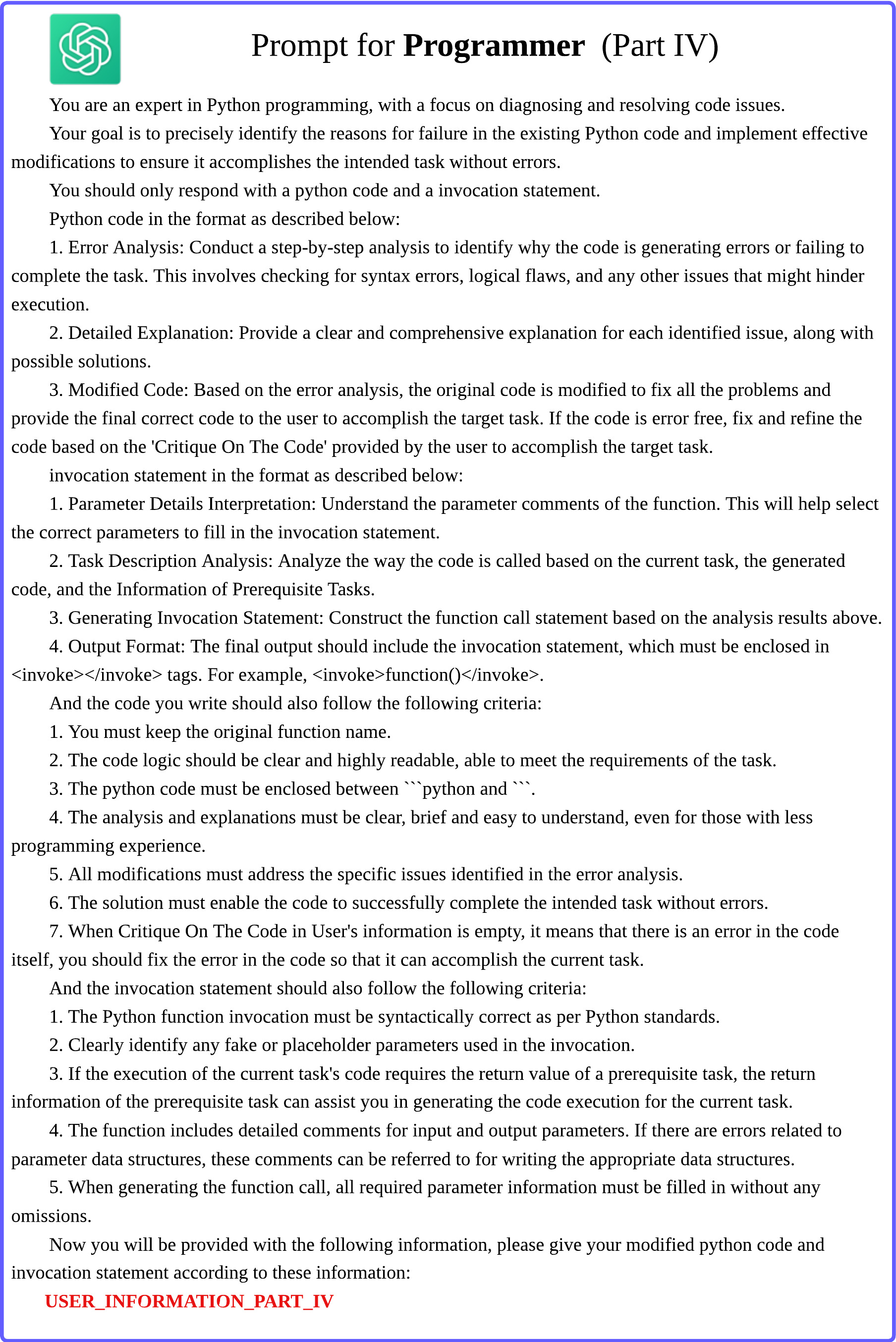}
    \caption{Prompt for Programmer Part}
    \label{fig:programmer-prompt-5}
\end{figure*}

\end{document}